\documentclass[lettersize,journal]{IEEEtran}
\usepackage{amsmath,amsfonts}
\usepackage[section]{placeins}
\usepackage{algorithmic}
\usepackage[ruled]{algorithm2e}
\usepackage{array}
\usepackage[caption=false,font=normalsize,labelfont=sf,textfont=sf]{subfig}
\usepackage{textcomp}
\usepackage{stfloats}
\usepackage{url}
\usepackage{verbatim}
\usepackage{graphicx}
\usepackage{cite}
\usepackage{booktabs}
\begin{document}
\bstctlcite{IEEEexample:BSTcontrol}
\title{
Dim Small Target Detection and Tracking: A Novel Method Based on Temporal Energy Selective Scaling and Trajectory Association
}

\author{Weihua Gao, Wenlong Niu, Wenlong Lu, Pengcheng Wang, Zhaoyuan Qi, Xiaodong Peng, Zhen Yang
\thanks{Manuscript received ; revised . (Corresponding author: Wenlong Niu.)}
\thanks{All authors are with the Key Laboratory of Electronics and Information Technology for Space Systems, National Space Science Center, Chinese Academy of Sciences, Beijing 100190, China. Weihua Gao is also with the School of Computer Science and Technology, University of Chinese Academy of Sciences, Beijing 100049, China. (e-mail:gaoweihua22@mails.ucas.ac.cn; niuwenlong@nssc.ac.cn; luwenlong21@mails.ucas.ac.cn; wangpengcheng20@mails.ucas.ac.cn; qizhaoyuan@hhu.edu.cn; Pxd@nssc.ac.cn; yangzhen@nssc.ac.cn). }
}

\markboth{IEEE TRANSACTIONS ON GEOSCIENCE AND REMOTE SENSING}%
{Shell \MakeLowercase{\textit{et al.}}: A Sample Article Using IEEEtran.cls for IEEE Journals}



\maketitle

\begin{abstract}
The detection and tracking of small targets in passive optical remote sensing (PORS) has broad applications. However, most of the previously proposed methods seldom utilize the abundant temporal features formed by target motion, resulting in poor detection and tracking performance for low signal-to-clutter ratio (SCR) targets. In this article, we analyze the difficulty based on spatial features and the feasibility based on temporal features of realizing effective detection. According to this analysis, we use a multi-frame as a detection unit and propose a detection method based on temporal energy selective scaling (TESS). Specifically, we investigated the composition of intensity temporal profiles (ITPs) formed by pixels on a multi-frame detection unit. For the target-present pixel, the target passing through the pixel will bring a weak transient disturbance on the ITP and introduce a change in the statistical properties of ITP. We use a well-designed function to amplify the transient disturbance, suppress the background and noise components, and output the trajectory of the target on the multi-frame detection unit. Subsequently, to solve the contradiction between the detection rate and the false alarm rate brought by the traditional threshold segmentation, we associate the temporal and spatial features of the output trajectory and propose a trajectory extraction method based on the 3D Hough transform. Finally, we model the trajectory of the target and propose a trajectory-based multi-target tracking method. Compared with the various state-of-the-art detection and tracking methods, experiments in multiple scenarios prove the superiority of our proposed methods.
\end{abstract}


\section{Introduction}
\IEEEPARstart{M}{oving} targets detection in passive optical remote sensing is widely used in many fields, including intelligent monitoring, active defense of spacecraft, and maritime target surveillance \cite{Wang_2021_Research, Dutta_2022_Extraction, Liu_2023_Single-Frame}. However, due to the limitations of the optical sensors, the imaging resolution is inversely related to the detection distance and the detection field of view. To detect the target as fast as possible, a long detection range and a large detection field of view are necessary, causing the target to occupy only a few pixels in the image, and lack of shape, color, and texture features. Dim targets and pervasive background interference result in a low SCR. Detecting the dim and small targets with low SCR in a complex background is a difficult problem. Moreover, tracking multiple targets is also an indispensable task for the early warning system. However, for weak and small targets, it is more challenging to utilize their scarce information to achieve tracking.

Visible and infrared cameras are commonly used for PORS. Visible light cameras typically offer higher spatial resolution and frame rate, and the movement of targets can be well recorded. But in low-light conditions or adverse weather, visible light cameras struggle to perform effectively. Infrared cameras, operating on the principle of thermal imaging from target heat emissions, offer advantages unaffected by lighting and weather conditions. However, they exhibit lower resolution and frame rates, resulting in suboptimal performance when imaging low-heat targets like small unmanned aerial vehicles (UAVs) \cite{Shi_2018_Anti-Drone,Fang_2022_Infrared}. Therefore, whether employing a visible light or infrared camera, addressing the challenge of detecting distant low SCR small targets is imperative.

Detection and tracking are two inextricably linked tasks, and their idea includes two categories: Track-Before-Detect (TBD) and Detect-Before-Track (DBT). Instead of detecting each image of sequences, TBD directly outputs the motion trajectory of the target. The variations among TBD series methods lie in the different strategies employed for trajectory searching. Common search strategies include dynamic programming \cite{Tian_2019_Novel}, matched filtering \cite{Javanmardi_2018_Visual}, multi-hypothesis testing \cite{Du_2022_DP-MHT-TBD}, particle filtering \cite{Tian_2022_Intelligent}. Although TBD series algorithms exhibit some detection efficacy for low SCR targets, their high search frequency and computational complexity make them impractical for real-time detection. Additionally, these algorithms need the target to have high saliency, which does not apply to small and weak targets. Therefore the algorithms based on TBD are not widely used due to those shortcomings.

The principle of the DBT method is to detect single-frame or multi-frame images first, get the candidate target, and then comprehensively utilize the motion and appearance information of the target to track the target, and finally output the correct trajectory. The algorithms based on DBT have been widely used due to their high interpretability and fast calculation speed, and consist of two steps: detection and tracking, which will be introduced separately.

A large number of small target detection algorithms have been proposed, which can be divided into two categories based on the information utilized: single-frame detection methods and multi-frame detection methods. The single-frame detection method suppresses the background and detects the target based on the characteristics of the target and the background. The spatial filtering (SF) based method considers that the background has a strong continuity, and the purpose of suppressing the background and highlighting the target is achieved by filtering out the clutter through a well-designed filter \cite{Zeng_2006_design,Zhao_2023_Fast}. The methods based on the human visual system (HVS) assume that the target has highly salient features relative to the background, as manifested by variance, intensity \cite{Chen_2014_Local,Wei_2016_Multiscalea,Deng_2016_Infrared,Chen_2023_Simplified}. While the low-rank and sparse decomposition (LRSD) based methods assume that the background is a low-rank component and the target has global sparsity, which is converted into a matrix decomposition problem to get the target image \cite{Gao_2013_Infrared,Dai_2017_Reweighted,Sun_2018_Infrared,Liu_2023_Single-Frame}. All of these methods place certain constraints on the target and the background, making it difficult to be effective for weak targets at long distances. Moreover, single-frame detection does not utilize the motion information of the target, and it is difficult to distinguish the target from the false target. Some methods use deep neural networks to detect weak targets, which are not applicable due to the lack of training data and high computational effort \cite{Dai_2021_Attentional,Nian_2023_Local,Wu_2023_UIU-Net}.

For moving small targets, spatial information is very scarce, using the spatiotemporal information from the target motion is a better choice. This is the advantage of the multi-frame detection method. Temporal difference is a straightforward idea, but it only utilizes adjacent frames, does not suppress the background well, and tends to suppress weak targets as background\cite{Sengar_2017_Moving}. Most of the existing multi-frame detection methods are carried out by fusing multi-frame information with single-frame detection methods and do not fully use the motion information of the target \cite{Li_2019_Weak,Du_2020_Infrareda,Pang_2022_Novel,Li_2023_Sparse}. However, some methods take the intensity temporal profile (ITP) formed over time by a single pixel as the study object. When a target passes to a pixel, a transient disturbance will be created in the ITP of the pixel. Thus the problem of small target detection in images is converted into a transient disturbance detection problem in ITP. Researchers proposed some ITP-based detection methods, and realized nice detection of low SCR small targets \cite{Silverman_1996_Temporal,  Tzannes_2002_Detecting,Liu_2007_Temporal,Bae_2010_Small,Liu_2015_Moving,Liu_2015_Temporal,Niu_2018_Moving,Wu_2018_Weak,ChenHao_2020_Dim,Liu_2022_Moving,Wang_2023_Dim}. The ITP-based method has the potential to solve the problem of detecting small targets with low SCR in complex scenes.

After the introduction of detection methods, we briefly describe the tracking algorithms for small targets. Target tracking is the process of consistently locating and monitoring a designated target within an image sequence or video frame. According to the different tracking processes, there are two mainstream schemes for target tracking: Tracking without Detection and Tracking while Detection. The former need to select a region of interest (ROI) as the target, and then only rely on the tracker to track the target region. The latter uses the detection results as inputs to the tracker and updates the target state during the tracking process. Most of the existing tracking algorithms are based on the spatial features of the target in a single frame of the image. Due to the lack of spatial features on weak and small targets, existing trackers will easily lose the correct position of a target, especially if multiple targets are approaching \cite{Wang_2020_Small,Wan_2022_Infrared,Kou_2023_Infrareda}.

In this work, we focus on the ITPs of pixels to achieve detection of target trajectories in a multi-frame detection unit and realize multi-target tracking based on the trajectories. Our main idea is to use an ITP-based detection method to obtain the original target trajectory, use a trajectory extraction method based on the 3D Hough transform to suppress false alarm points in the background, and finally use trajectory-based tracking methods to realize multi-target tracking. The main contributions of our work are as follows:

\begin{enumerate}
    \item We analyze the difficulty of detecting small targets with low SCR based on single-frame and the feasibility of detection based on multi-frame under long-range and large field-of-view conditions. According to the analysis, we have fully investigated the statistical properties of the target, background, and noise components in the ITP using multiple frames as the detection unit. Based on the statistical property, we design a temporal selective scaling function to detect the transient disturbance formed by the target. Our detection method is parallelized and extremely easy to be accelerated by GPUs.
    \item To solve the contradiction between the detection rate and the false alarm rate brought by the traditional threshold segmentation, we associate the spatio-temporal characteristics of the target trajectory and propose a trajectory extraction method based on the 3D Hough transform, which can suppress a large number of false alarms without losing the detection rate.
    \item To solve the problem that small targets in low SCR are difficult to track effectively, we take the target trajectory generated by the detection method and the trajectory extraction method in the multi-frame detection unit as the tracking object, model the state of the target trajectory, design the cost matching function, and improve the existing tracking strategy by combining with the characteristics of the trajectory, propose our multi-small target tracking method based on the trajectory.

\end{enumerate}

The remaining sections of this article are structured as follows. In Section II, we introduce some related works about small target detection and tracking. In Section III, we give some necessary mathematical symbols and definitions. We analyze the disadvantages of single-frame-based detection and the advantages of multi-frame-based detection. The proposed detection method based on temporal energy selective scaling is given in detail. After the detection method, we present a trajectory extraction method based on the 3D Hough transform. Following trajectory extraction, the tracking method for multiple small targets with low SCR based on the trajectory is given in detail. In Section IV, a series of experimental details and method performance are analyzed. Finally, we conclude with a summary of the entire article.

\section{Related Works}
In this section, we will first introduce the detection methods based on ITP. Subsequently, we present some tracking methods.
\subsection{ITP-based Detection Methods}
Unlike traditional detection methods that rely on a single frame as the detection object, ITP-based methods utilize the intensity temporal profile created by a single pixel over time. As the target traverses multiple pixels within the image sequence, a transient disturbance manifests on the ITP generated by the pixels traversed by the target. Detecting small targets in images can be converted to detecting transient disturbance (target signal) in ITP.

Research on ITP-based methods has a history of several decades. Silverman et al. \cite{Silverman_1996_Temporal} were the first to carry out related research and proposed a composite triple temporal filter (TTF) for suppressing cloud clutter and extracting the target signal. Tzannes et al. have done a lot of work. They first proposed a multiscale temporal filtering algorithm using wavelet transform in 1997 \cite{Tzannes_1997_Temporal}. Followed by mathematical modeling of the ITP, a hypothesis-testing algorithm \cite{Tzannes_2002_Detecting} was proposed, which improves the detection of the target signal in the ITP. Liu et al. proposed a filtering algorithm based on the connecting line of the stagnation points (CLSP) to obtain the target signal by subtracting the CLSP baseline of the background ITP from the original ITP \cite{Liu_2007_Temporal}. Tae-Wuk et al. used a cross product of temporal pixels based on ITP to obtain the background ITP and subsequently differenced to obtain the target signal to achieve detection \cite{Bae_2010_Small}. Liu et al. used a nonlinear filter to filter out clutter and discriminate the presence or absence of a target signal based on the variance of different segments of the ITP \cite{Liu_2015_Moving}. Niu et al. also did a lot of work on this problem. They first studied the higher-order spectrum of ITP and found that there is a difference between the target signal and the background ITP, and proposed a detection algorithm\cite{Niu_2018_Moving}. Subsequently, they proposed a kernel function method based on statistical distance, which achieved good detection results \cite{Wu_2018_Weak}. Excitingly, they found that the improvement of the frame rate can help to improve the detection effect. Hao et al. used local variance and global variance as indicators to detect the target signal \cite{ChenHao_2020_Dim}. Liu et al. proposed a detection method based on the similarity of ITP in neighboring pixels \cite{Liu_2022_Moving}. Wang et al. used a 1D CNN to learn the background signal of ITP and obtained the target signal by differencing to improve the detection\cite{Wang_2023_Dim}. 

These methods are effective for targets with high SCR, however, when the SCR is reduced, the target signal is submerged in the background and the detection performance of these methods is greatly declined. Therefore, it is necessary to fully study the background component, noise component, and target component in ITP, to propose a selective scaling detection method that can well suppress the background and enhance the target.

\subsection{Tracking Methods}
It was mentioned earlier that tracking methods can be categorized into tracking without detection and tracking while detection. These two types of methods are described below.
\subsubsection{Tracking Without Detection}
Tracking without detection methods include Meanshift, Particle Filter, and Kernel Correlation Filter Tracking. The Meanshift tracking method is a probability density distribution-based method. Mean shift starts by modeling the target region to be tracked, and the search for the target will continue in the direction of the increasing probability gradient until several iterations converge to the location of the densest local peak of the probability density distribution\cite{Yu_2018_Moving}. The particle filter tracking method is a method based on particle distribution statistics\cite{Wan_2022_Infrared}. Firstly, the tracked target area is modeled and a similarity measure is defined to measure the degree of match between the particles and the target; then in the process of target search, some particles are scattered around according to a certain distribution; finally, the possible location of the target is determined by measuring the similarity of different particles. The tracking algorithm based on kernel correlation filtering introduces the method of correlation filtering to target tracking, where the filtering template is correlated with the target candidate region, and the location of the maximum output response is the target location. Common methods include MOSSE\cite{Bolme_2010_Visual}, CSK\cite{Henriques_2012_Exploiting}, KCF\cite{Henriques_2015_High-Speed}, ECO\cite{Danelljan_2017_ECO}, CSRT\cite{Lukezic_2017_Discriminative}. The MOSSE is the first correlation filter tracking method, which minimizes the mean squared error as the objective function and uses multiple samples of the target as the training set to generate a better filter. The CSK method employs a kernel tracking method based on a cyclic matrix and solves the sample redundancy problem caused by the sparse sampling of the MOSSE method. After that, scholars have proposed a series of correlation filtering methods such as KCF, ECO, and so on. In recent years, some deep learning based target tracking methods have been proposed\cite{Wang_2022_Low-Slow-Small,Kou_2023_Infrareda,Yang_2023_SiamMDM}, representative of which are the Siamese networks. However, those networks are difficult to apply to small targets that lack features.
The tracking without detection methods rely on manual input of the target position and does not have the capability of autonomous discovery and tracking.

\subsubsection{Tracking While Detection}
Common trackers include Kalman filter\cite{Gao_2023_Research}, SORT\cite{Bewley_2016_Simple} and DeepSORT\cite{Wojke_2017_Simple}. The Kalman filter is a classical predictive tracking algorithm, which is an algorithm that utilizes the linear system state equation to optimally estimate the system state from the system input observations. SORT is a multi-target tracking method proposed in 2016, which mainly utilizes the Kalman filter and Hungarian algorithm to realize multi-target tracking and greatly improves performance. SORT has three main steps: prediction, association, and update. When determining the position of the target in the current frame, it is necessary to predict the position of the tracked object in the current frame, then make the association between the targets according to the target detection result and the prediction result of the tracker, and finally update and maintain the state of the tracker according to the association result. DeepSORT is based on the framework of SORT, with the addition of new cascade matching and deep appearance information. DeepSORT solves the problem of missed target detection and track confirmation through the addition of trajectory state and trajectory confirmation, which further improves the target tracking capability and meets the evaluation index of SOTA.

In this article, we use tracking while detection method, but the existing methods make it easy to track the incorrect target. So improving the tracking method by combining our detection method is necessary. 

\section{Proposed Detection and Tracking Method}
\begin{figure*}[!t]
\centering
\includegraphics[width=6in]{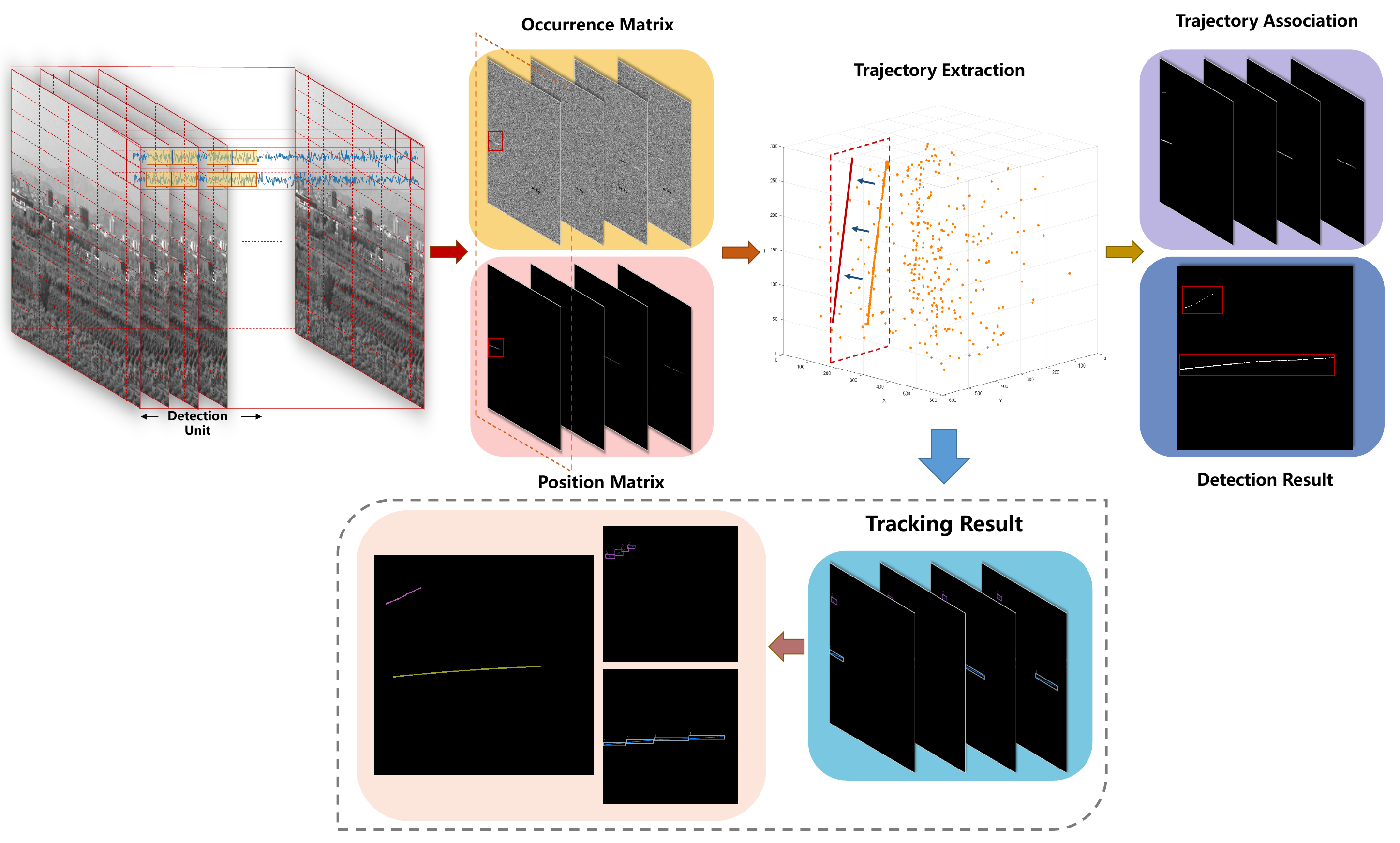}
\caption{Framework of our proposed detection and tracking method. First, we compute the distance of each ITP in a multi-frame detection unit in a parallel way, obtaining the trajectory position matrix and the occurrence moment matrix. Then, we perform trajectory extraction based on the 3D Hough transform to eliminate false alarm points. After the trajectory extraction, we can directly associate the trajectories to get the final detection results. It is also possible to track the trajectories of the detection unit to get the trajectory tracking result.}
\label{detection_tracking_fig}
\end{figure*}

In this part, we analyze the shortcomings of the single-frame detection method and the advantages of the multi-frame detection method for low SCR small targets. Subsequently, we propose our detection and tracking method, which is shown in Fig. \ref{detection_tracking_fig}.

\subsection{Analysis of Single-Frame and Multi-Frame Detection}
We first demonstrate the disadvantages of the single-frame detection method in detecting dim small targets in terms of target size, target intensity, and SCR. Subsequently, we analyze the SCR gain after the accumulation of multiple frames.

\begin{figure}[h]
  \includegraphics[width=3.5in]{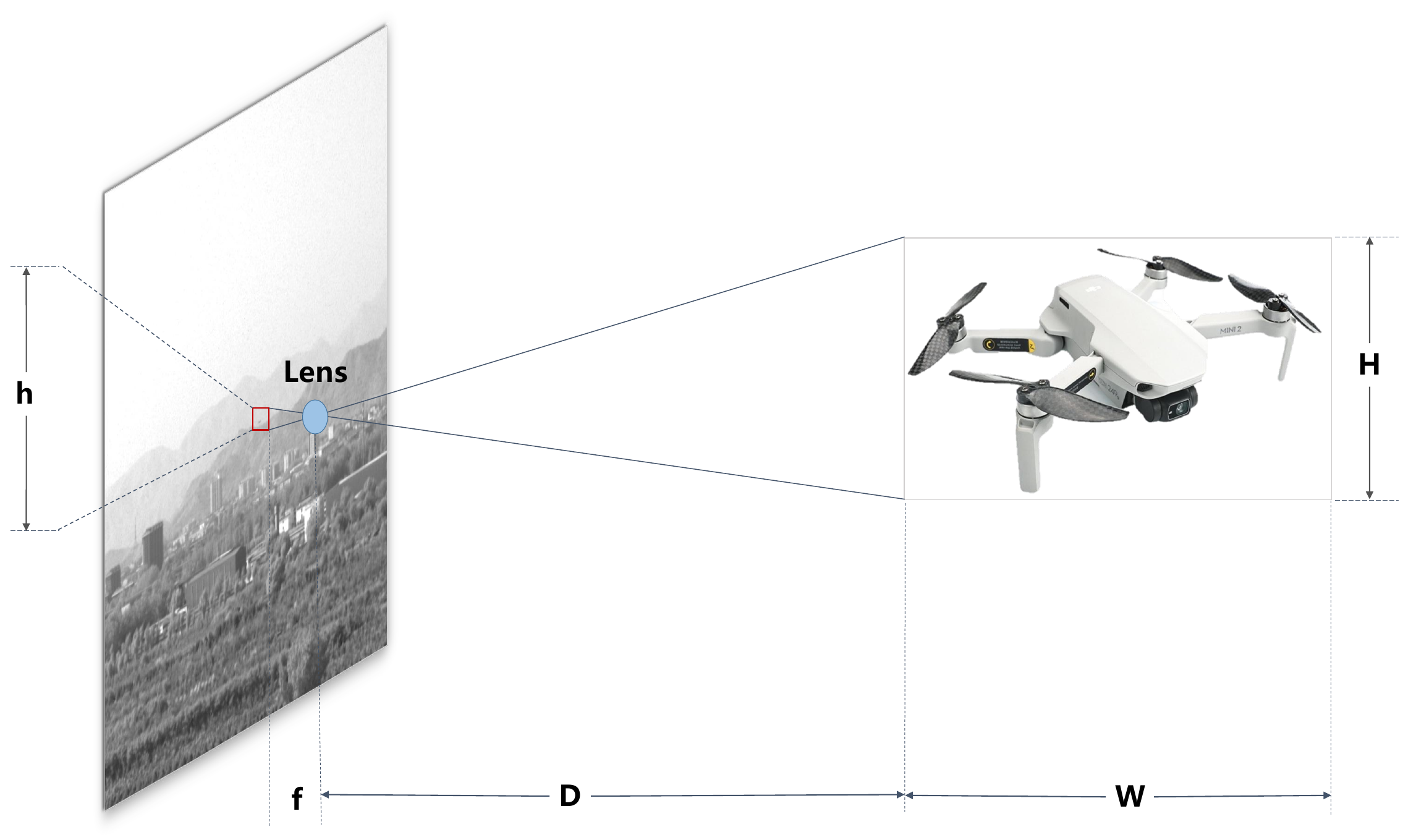}
\caption{Principles of Optical Imaging.}
\label{obj_img_eqa_fig}
\end{figure}

The simplified  principle of optical imaging is shown in Fig.\ref{obj_img_eqa_fig}. Under the principle of pinhole imaging, the object-image size relationship is shown in Equation \ref{obj_img_rel_equ}:
\begin{equation}
h=\frac{f}{D} \times H
\label{obj_img_rel_equ}
\end{equation}
where $f$ is the focal length of the lens, $D$ is the imaging distance, and $H$ is the actual size of the target. 

It can be seen that the size of the target after imaging is proportional to the focal length of the lens and inversely proportional to the imaging distance. To ensure the earliest possible detection of the target, it is necessary to maximize both the detection distance and the field of view for detection. However, as the detection distance $D$ increases, the size $h$ of the target after imaging becomes smaller. Small targets at long distances occupy only a few pixels in the image and lose shape, texture and size information, cannot be detected by single-frame detection methods based on the target's spatial features.

Next, consider the relationship between target intensity and detection distance. The intensity after imaging is determined by the number of photons received by the image element, and the photoelectric converter within the image element converts the optical signal into a digital signal, which is processed through signal amplification, correlation double sampling, and other processes to obtain an intensity value. The simplified photo-intensity conversion equation is as follows:
\begin{equation}
H(a,b)=k\int^{t+\Delta t}_t dt \iint^{R_D}_{(x,y)}E(x,y,t)dxdy
\label{photo-intensity_conv_equ}
\end{equation}

$E(x,y,t)$ denotes the photon number at point (x,y) at moment t, k is the conversion efficiency, $R_D$ denotes the imaging region of the pixel $(a,b)$, and $\Delta t$ is the exposure time. As the imaging distance becomes farther, the imaging region $R_D$ of a pixel will become larger, which means that the target and the background will be fused, and the high-frequency part of the target will be lost firstly until the range of the target $R_T$ is less than the imaging region of a pixel ($R_T \leq R_D$), the target will occupy sub-pixel in the image. The intensity of the target will be fused with the background continuously. In this condition, the dim target cannot be detected by previous single-frame detection methods based on target intensity.

We use SCR to describe this process. The equation of SCR is shown as follows:
\begin{equation}
\text{SCR}=\frac{|\mu_T-\mu_B|}{\sigma_B}
\label{SCR_equ}
\end{equation}

$\mu_T$ denotes the mean intensity of the target, $\mu_B$ and $\sigma_B$ are the mean intensity and the standard deviation of the background.

As the imaging distance increases, the process of target degradation over the space and fusion with the background can be viewed as a process of pixel mixing. We assume that the target degrades from M × K size to one pixel. The spatial signal of the target is expressed as $T(i,j)$, $(i,j)$ denotes the pixel position, and there are a total of N pixels in the M × K range. The equation for the SCR is rewritten as:
\begin{equation}
\text{SCR}=\frac{|\mu_T-\mu_B|}{\sigma_B}=\frac{|\sum T(i,j)/N-\mu_B|}{\sigma_B}
\label{reSCR_equ}
\end{equation}

When the target is degraded to one pixel, the target intensity of the pixel is:
\begin{equation}
A_s=a*\sum{\frac{T(i,j)}{N}}=a*\mu_T,\ \ \ \  0<a\leq 1
\label{Amixed_equ}
\end{equation}

where $a$ is the fusion factor between the target and the background, a variable related to the size and shape of the target. The more irregular the shape and the larger the size of the target, the larger the $a$.

The noise component is analyzed there. The composition of noise $N$ is shown as equation \ref{noise_composi_equ}
\begin{equation}
N=N_p+N_{read}+N_q
\label{noise_composi_equ}
\end{equation}

$N_p$ is the shot noise obeying Poisson distribution, $N_{read}$ is the readout noise including thermal and $1/ f$ noise, and $N_q$ is the quantization noise obeying uniform distribution. When the imaging distance increases, the number of photons that should be received by a total of M $\times$ K pixels is concentrated into a single pixel. The increase in the number of photons leads to an increase in the shot noise; the rest of the noise will not be affected approximately. Therefore the standard deviation of the noise becomes larger ($\sigma_{(B,L)} \geq \sigma_B$), while the average intensity of the background ($\mu_B$) can be considered approximately unchanged, and the ratio of the SCR is as follows:
\begin{equation}
\frac{\text{SCR}_\text{L}}{\text{SCR}}=|\frac{a*\mu_T-\mu_B}{\mu_T-\mu_B}|*\frac{\sigma_B}{\sigma_{B,L}}\leq 1
\label{SCR_ratio_equ}
\end{equation}
where $\text{SCR}_\text{L}$ denotes the degraded SCR.

\begin{figure}[h]
\centering
  \includegraphics[width=2.5in]{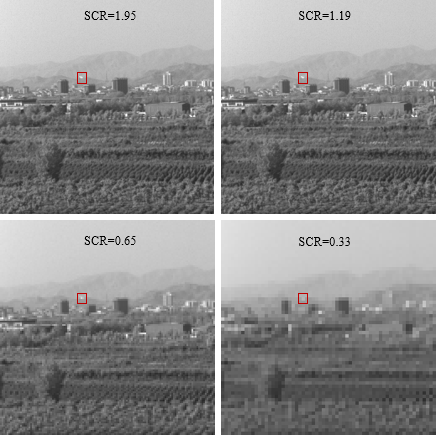}
\caption{SCR reduction after doubling, quadrupling, and octupling the detection distance.}
\label{down_SCR_fig}
\end{figure}

This indicates that the SCR decreases as the detection distance increases. Fig.\ref{down_SCR_fig} briefly illustrates this process.

As can be seen, both from the perspective of the imaging principle and the SCR, the detection methods based on single-frame are difficult to apply to such dim and small targets. By fully sampling the target motion process, we can accumulate the energy of the weak target in consecutive frames to increase the SCR and contribute to improving detection performance. Again, we use SCR to describe this process.

We can approximately model\cite{Gao_2013_Infrared} the target-prensent image as:
\begin{equation}
f_I=f_B+f_N+f_T
\label{image_model_equ}
\end{equation}

where $f_B$ denotes the background component, $f_N$ denotes the noise component, and $f_T$ denotes the target component. There are various sources of noise in the image, and according to the central limit theorem, the total noise can be considered to obey a Gaussian distribution, $G_{N} \sim N(0,\sigma_N^2)$.

The following discussions are all with the same exposure time. Assuming that the velocity of the moving target is $V$, the imaging region of a single pixel is $S\times S$, and the frame rate of the detector is $F$, the number of frames that the target stays in the single pixel is $K=S/V*F$. Accumulating the energy of K frames. Since the noise components of each frame are independent and the sum of the Gaussian distribution still obeys the Gaussian distribution, the accumulated noise obeys a distribution of $G_{N,K} \sim N(0,K\sigma_N^2)$. If the intensity of the target is $A$, without considering the point spread effect, the intensity value of the target after accumulating K frames is $A_K=KA$, and the SCR gain is:
\begin{equation}
\text{SCR}_{\text{gain}}=\frac{\text{SCR}_\text{K}}{\text{SCR}}=\frac{\frac{|KA-K\mu_B|}{\sqrt{K}\sigma_N}}{\frac{|A-\mu_B|}{\sigma_N}}=\sqrt{K}
\label{SCR_Kgain_equ}
\end{equation}
where $\text{SCR}_\text{K}$ is the SCR after accumulating K frames.

Apparently, as $K$ increases, the SCR of a small target increases. Under the premise of constant target motion speed, $K$ depends on the detector frame rate, so the higher the frame rate, the easier the small target can be detected.

Based on the above analysis, we found that the single-frame detection method is defective in dim small target detection, while the multi-frame detection method can reduce the detection difficulty. Therefore, we carry out the research of multi-frame detection method. Our detection method based on temporal energy selective scaling (TESS) is presented next.

\subsection{Temporal Energy Selective Scaling Detection Method}
The ITP of a pixel consists of three components: the background signal, the noise signal, and the target signal. Whether the target passes through this pixel is a hypothesis-testing problem:

\begin{align}
H_0: I_{(x,y)}(k)&=b_{(x,y)}(k)+n_{(x,y)}(k) \nonumber \\
H_1: I_{(x,y)}(k)&=b_{(x,y)}(k)+n_{(x,y)}(k)+t_{(x,y)}(k) 
\label{H_equ}
\end{align}

$H_0$ is the hypothesis that the target did not pass through the pixel, and $H_1$ indicates that the target passed through the pixel. The state of each pixel is $H_0$ or $H_1$. If we know all pixel states, the small target detection problem will be solved. To solve for the state of the pixel, we must know the properties of the components in the ITP.

\begin{figure}[h]
\centering
  \includegraphics[width=3.5in]{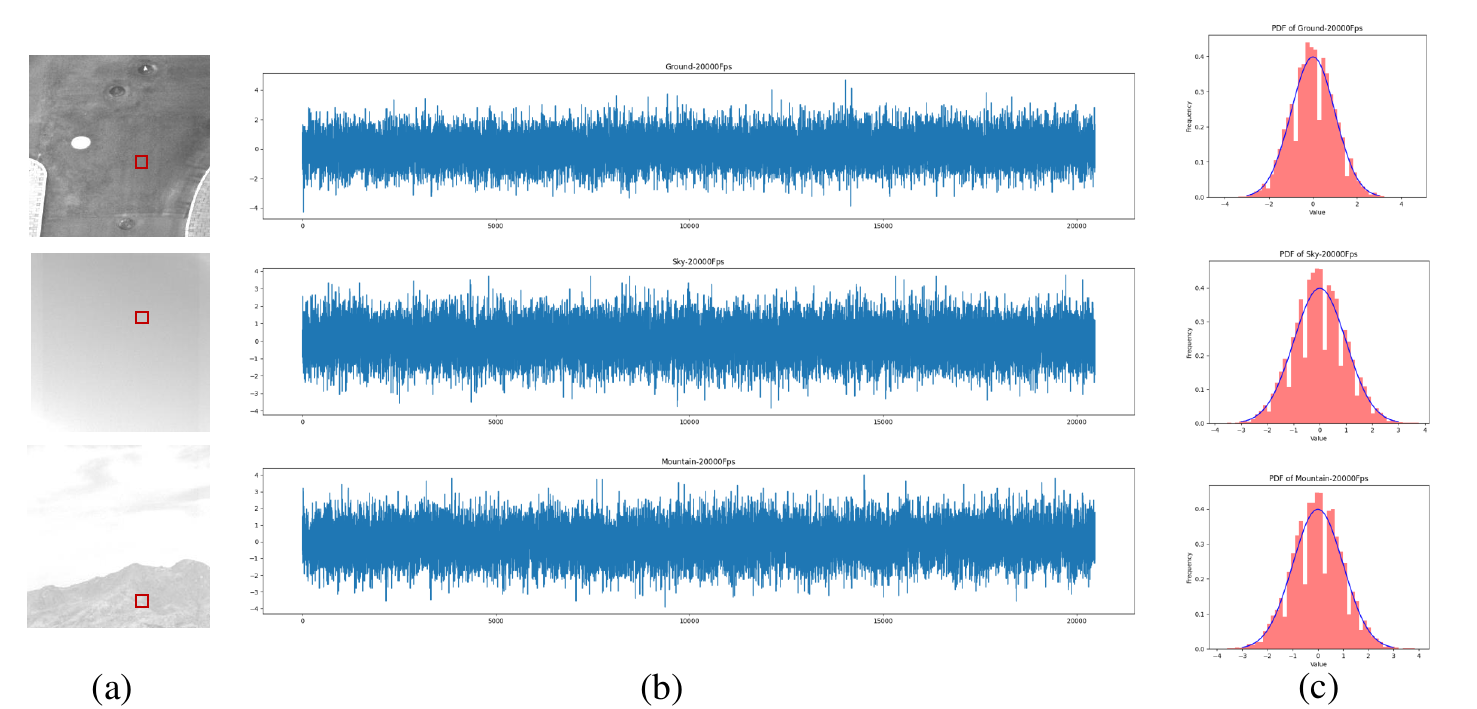}
\caption{Probability density functions (PDF) of typical pixel for different scenes. (a) Different scenes. (b) Noise signal of a typical pixel. (c) PDF of noise signals.}
\label{PDF_fig}
\end{figure}

Under gaze imaging conditions, the signal in a static background is a short-time smooth random signal that can be described by a constant $C$. According to the previous section, the noise signal obeys a Gaussian distribution, $G_{N} \sim N(0,\sigma_N^2)$. We also verified this conclusion. We selected typical pixels in different scenes and analyzed the probability density function of their ITP, as shown in Fig \ref{PDF_fig}.

\begin{figure}[h]
\centering
\includegraphics[width=3in]{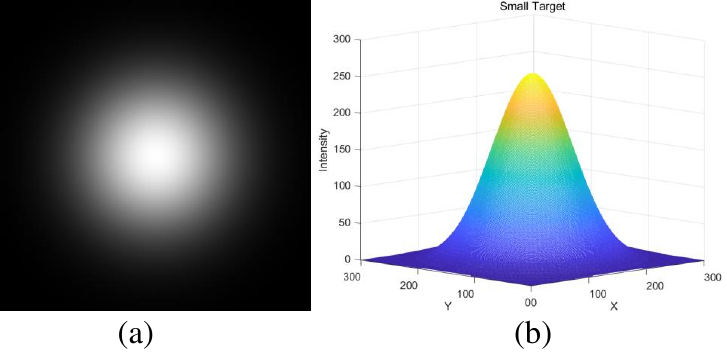}
\caption{The 2D and 3D simulations of small target. (a) The 2D simulation of a small target (b) The 3D simulation of a small target.}
\label{target_fig}
\end{figure}

The formation of the target signal is more complex and is determined by the shape of the small target in space. The small target at long distance can be approximated as a point and described using a point spread function:
\begin{equation}
\text{PSF(x,y)}=A\text{exp}\left(-\left(\frac{x^2}{2\sigma_x^2}+\frac{y^2}{2\sigma_y^2}\right)\right)
\label{PSF_equ}
\end{equation}

where $\sigma_x$ is the diffusion coefficient in the horizontal direction and $\sigma_y$ is the diffusion coefficient in the vertical direction, which are related to the target velocity. The 3D diagram of the small target is shown as Fig. \ref{target_fig}. 

The target signal formed as the target passes through the pixel approximates a bell-shaped signal, that is, the result of a tangent to the target's 3D map in Fig\ref{target_fig}(b) in it's moving direction and stretched.

\begin{figure}[h]
\includegraphics[width=3.5in]{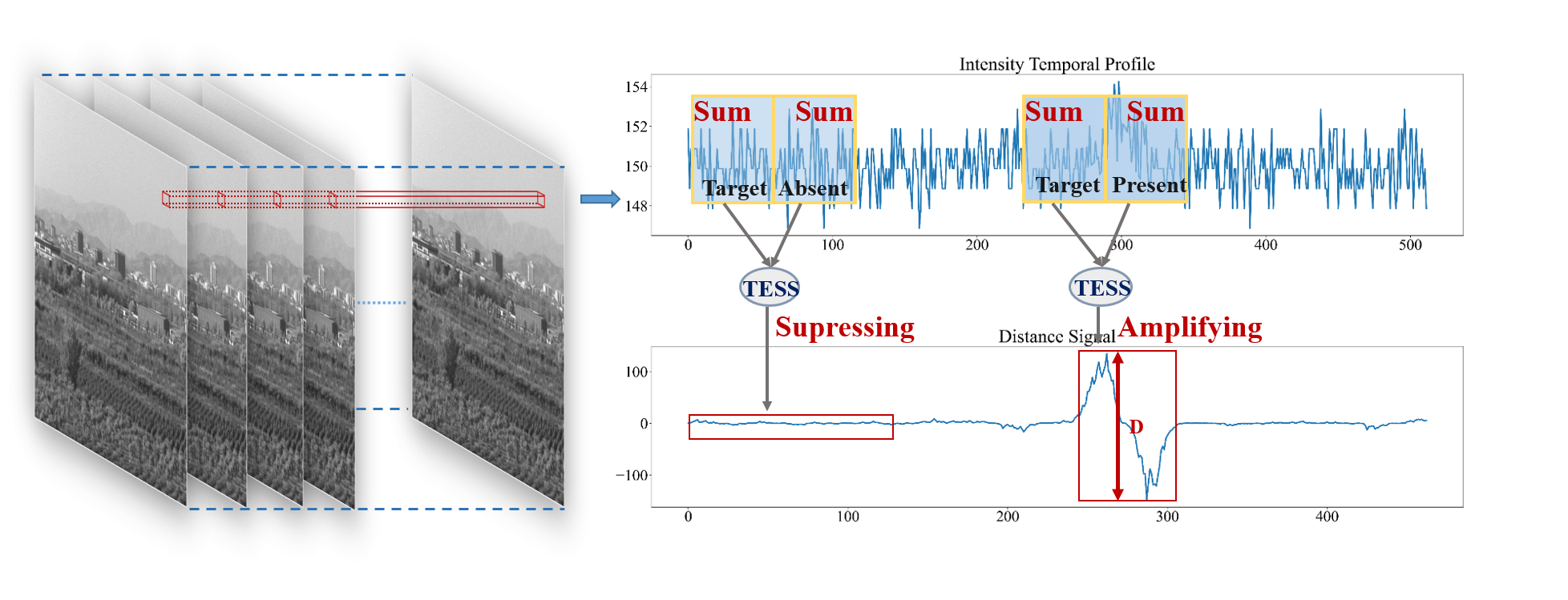}
\caption{The detection schematic of our method.}
\label{detection_schematic_fig}
\end{figure}

We already know the properties of each component in the ITP, so we can propose a detection method based on the difference of these properties. Our detection schematic is shown in Fig \ref{detection_schematic_fig}. 

The TESS is our detection method based on temporal energy selective scaling. We can see that the background and noise are effectively suppressed and the target is greatly amplified. The specific scaling principle is to use a sliding window to calculate the distance between the front half ($W_f$) and the rear half ($W_r$) inside the window. The distance calculation formula is as follows:

\begin{equation}
\text{D}=\left[\sum^{W_r}\text{ITP}(i)-\sum^{W_f}\text{ITP}(j) \right]\left(e^{|\mu_r-\mu_f|}-1\right)
\label{TESS_equ}
\end{equation}

where $\mu_f$ and $\mu_r$ are the average intensity of $W_f$ and $W_r$. This equation consists of two parts, the first is the difference in temporal accumulated energy and the second is the exponential selective scaling. We first demonstrate the effectiveness of the front part for detecting the target signal. The process of subtracting the front and rear windows is as equation \ref{subtract_equ}.

\begin{eqnarray}
    \text{D}' & = &\sum^{W_r}\text{ITP}(i)-\sum^{W_f}\text{ITP}(j) \nonumber \\
             & = & \sum^{W_r} \left[b_{(x,y)}(i)\right]-\sum^{W_f} \left[b_{(x,y)}(j)\right] \nonumber\\
             & + & \sum^{W_r} \left[n_{(x,y)}(i)\right]-\sum^{W_f} \left[n_{(x,y)}(j)\right] \nonumber\\
             & + & \sum^{W_r} \left[t_{(x,y)}(i)\right]-\sum^{W_f} \left[t_{(x,y)}(j)\right]
    \label{subtract_equ}
\end{eqnarray}

Since the background signal is a short-time smooth random signal. The noise signal obeys a Gaussian distribution, and the sum of the Gaussian distributions still obeys the Gaussian distribution,$\sum G_N \sim (0,\sum \sigma^2)$. So, the ideal result of equation \ref{subtract_equ} is:

\begin{eqnarray}
    \text{D}' & = & \sum^{W_r} \left[b_{(x,y)}(i)\right]-\sum^{W_f} \left[b_{(x,y)}(j)\right] \nonumber\\
             & + & \sum^{W_r} \left[n_{(x,y)}(i)\right]-\sum^{W_f} \left[n_{(x,y)}(j)\right] \nonumber\\
             & + & \sum^{W_r} \left[t_{(x,y)}(i)\right]-\sum^{W_f} \left[t_{(x,y)}(j)\right] \nonumber\\
             & = & \sum^{W_r} \left[t_{(x,y)}(i)\right]-\sum^{W_f} \left[t_{(x,y)}(j)\right] +\varepsilon 
    \label{subtract_result_equ}
\end{eqnarray}

Where, the $\varepsilon=0$ when the sliding window tends to infinity in the static background ($\lim\limits_{W \to \infty} \varepsilon =0$). Therefore, $D'$ is theoretically determined by the target signal only. When there is no target signal in the window, $D' = 0$. When the window gradually slides over the target signal, $D$ increases and then decreases. When the target signals all appear in the rear half window or front half window, $D$ gets the maximum or minimum, as Fig \ref{detection_schematic_fig} shown. 

However, this is only a mathematical derivation of the ideal case. In practice, due to lens jitter or clutter interference, the background cannot be completely static, and the assumption of short-term smoothness is not satisfied. Some scholars use Markov models to describe the changing background \cite{Liu_2015_Temporal}, but there is no use for detection. Instead of the Markov model of background, we use exponential selective scaling to suppress background changes.

\begin{figure}[h]
\includegraphics[width=3.5in]{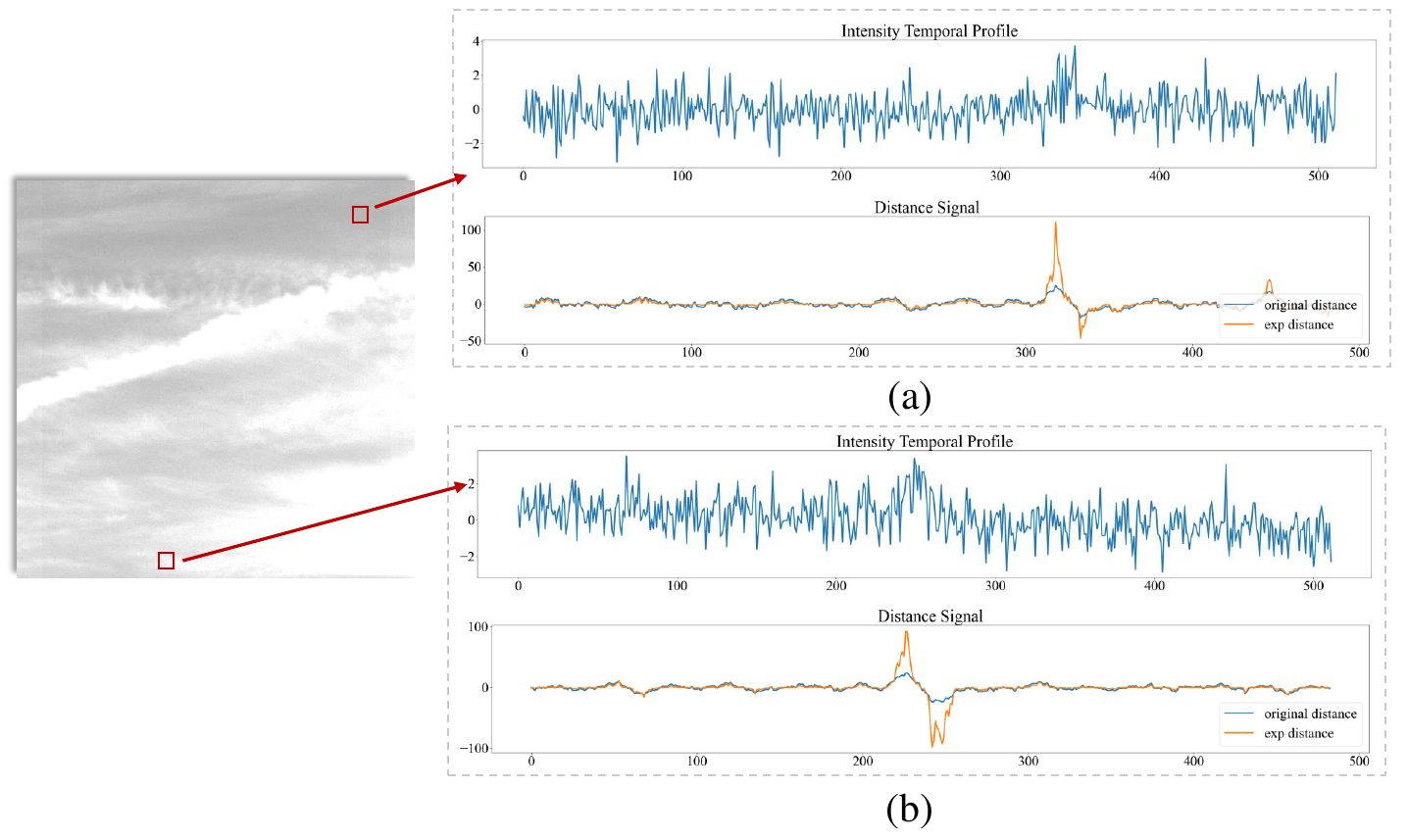}
\caption{The distance signals of different ITPs. (a) The signals of a pixel in a static background. (a) The signals of a pixel in cloud edge.}
\label{ITPs_distance_fig}
\end{figure}

We use the average intensity difference between the front and rear windows as the variable of the exponential function $\left(e^{|\mu_r-\mu_f|}\right)$. However, the background-to-background difference would also be amplified if only the exponential part was used , so we shifted the exponential function down one unit along the y-axis $\left(e^{|\mu_r-\mu_f|}-1\right)$. In this way, the smaller background distances are suppressed and larger target distances are amplified. To verify the effect of exponential selective scaling, we selected two ITPs to get the distance signal, one from a pixel with a stable background and the other from a pixel of cloud edge with a slowly changing intensity, as shown in Fig. \ref{ITPs_distance_fig}.

In the figure, the exp distance signal (orange line) is generated by the exponential selective scaling, while the original distance signal (blue line) does not use it. Obviously, after exponential selective scaling, the target distance is amplified and the background distance is suppressed. The exponential module suppresses the background well even in ITP with intensity slowly changing.

After performing a primary sliding window on the ITP to obtain the distance signal, we perform a secondary sliding window on the distance signal to obtain the maximum distance and the moment of its occurrence. We utilize the maximum distance and occurrence moment as the posterior intensity of the pixel and the time at which the target passes through the pixel. The size of the secondary sliding window is $W_r+W_f$. After sliding the window twice and a simple thresholding, we obtained the position and occurrence matrix of target trajectories. The detailed detection process is shown in Algorithm \ref{detection_method}.

\begin{algorithm}
\caption{Detection of Small Target Trajectories in Multi-Frame Unit}\label{detection_method}
\KwIn{Multi-frame units $I$ containing $N$ images. Size of sliding window $W$. Detection threshold $\theta$.}
\KwResult{Trajectories position matrix: $P$ and occurrence time matrix: $T$.}
$P=0$, $T=0$, $L_D=\emptyset$\\
$W_h=W/2$, $i=0$\\
Standard $I$ along the time direction: $I=Standard(I)$\\
\While{$i+W \leq N$} {$W_1=I[i:i+W_h]$, $W_2=I[i+W_h:i+W]$\\
$E=Sum(W_1)-Sum(W_2)$\\
$\mu=E/W$\\
$D=E*\left(e^{|\mu|}-1\right)$\\
$L_D[i]=D$\\
$i=i+1$\\
\If{$i\geq W$} {
$W_S=L_D[i-W:i]$\\
$D=Max(W_s)-Min(W_s)$\\ 
$index=Where((P-D)<0)$\\
\If{$Len(index)>0$}{
$P[index]=D[index]$\\
$T[index]=i-W$
}
}
}
$P[P<\theta]=0$, $P[P\geq\theta]=255$\\
return $P,T$
\end{algorithm}

\subsection{Trajectory Extraction Method Based on 3D Hough Transform}
The process of thresholding detection results introduces a contradiction between the detection rate and the false alarm rate. We can lower the threshold to get a higher detection rate, but this will also bring more false alarm points. To address this issue, we fuse the position matrix and occurrence matrix into 3D space to realize trajectory extraction based on the 3D Hough transform, thus eliminating false alarm points. We first analyze the difference between target trajectories and false alarm points.

\begin{figure}[h]
\centering
\includegraphics[width=3.5in]{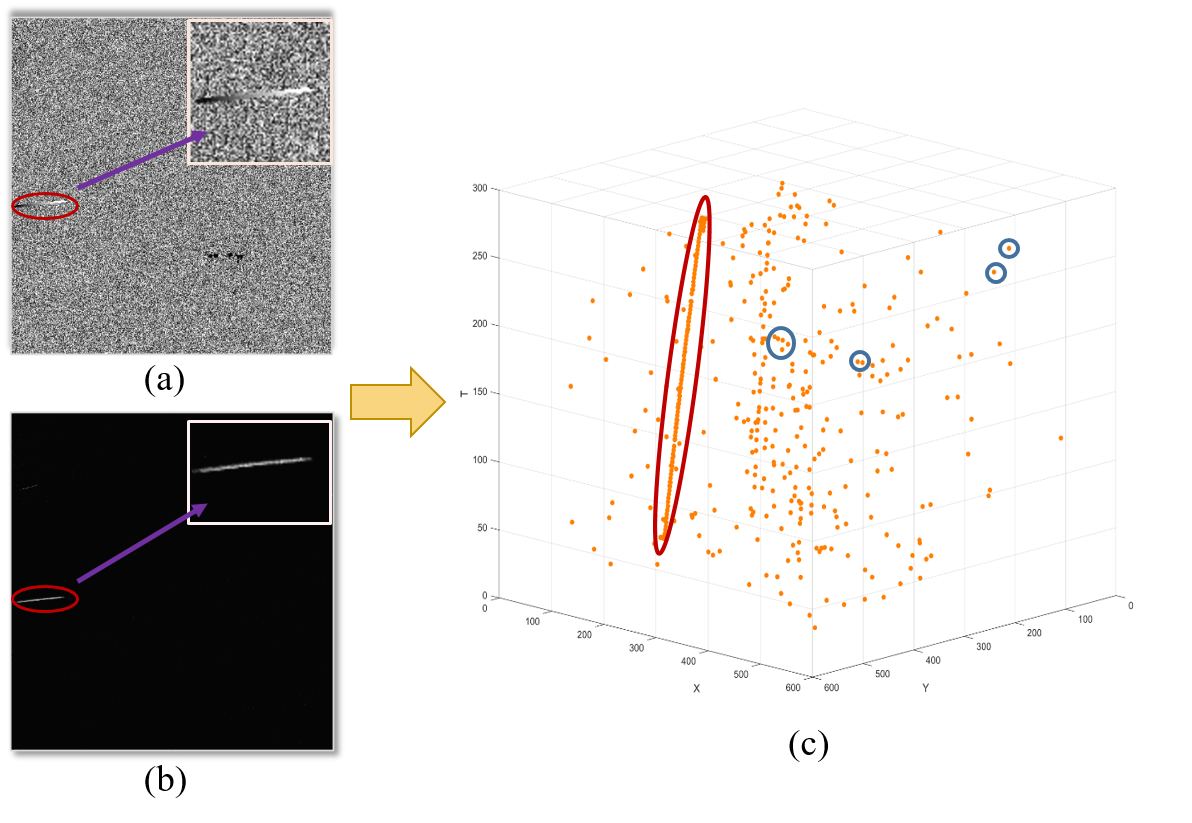}
\caption{The three-dimensional view of the detection result. (a) The occurrence matrix of the detection result. (b) The position matrix of the detection result. (c) The 3D map of the detection result.}
\label{3D_trace}
\end{figure}

Fig. \ref{3D_trace} illustrates the 3D map formed by the fusion of the position matrix and the occurrence matrix. The threshold of this detection result is 0, so there are a lot of false alarm points. The red circle is the 3D trajectory of the target, and the blue circle is some of the false alarm points. It can be seen that the 3D trajectory of the target is a line, while the false alarm points are discrete. This is due to the movement of the target is continuous in time and space. In space, the target moves over neighboring pixels. In time, the occurrence moments of the target are sequential. In Fig. \ref{3D_trace} (a), the higher intensity indicates the later occurrence moment of the target. While the false alarm points are discrete in space and are not sequential in time. Therefore we can detect the target trajectory as a line.

In a multi-frame detection unit, the target trajectory is a line with spatio-temporal correlation. The Hough transform is a well-established line detection method that can detect both straight and curved lines. So we detect the target trajectory in 3D space based on 3D Hough transform \cite{Jeltsch_2016_Hough}.

The main idea of the 3D Hough transform is to transform all points in 3D space to another parameter space, and the position of the larger votes in this parameter space represents a line in 3D space. How to find a parameter space with high accuracy and low computational complexity is a key issue.

The equation of a line $L$ in space can be described by equation \ref{line_equ}. 
\begin{eqnarray}
    p=a+t\hat{b}
    \label{line_equ}
\end{eqnarray}

where $p=(p_x,p_y,p_z)$ denotes any point, $a=(a_x,a_y,a_z)$ denotes the anchor point, $\hat{b}=(b_x,b_y,b_z)$ denotes the direction vector, and $||\hat{b}||=1$, $t\in (-\infty, \infty)$. When given an anchor point $a$ and a direction vector $\hat{b}$, the only line is determined, which requires 6 parameters. But one of the three parameters of $a$ is redundant actually, we can use Robert line representation to eliminate it \cite{Roberts_1988_new}.

$\beta$ is a plane passing through the origin and perpendicular to $L$, $(x',y')$ is the intersection of $\beta$ and $L$, are coordinates in the $\beta$ coordinate system. The equations for $x'$ and $y'$ are shown in equation \ref{xy_equ}.

\begin{equation}
\left\{\begin{array}{l}
x^{\prime}=\left(1-\frac{b_x^2}{1+b_z}\right) p_x-\left(\frac{b_x b_y}{1+b_z}\right) p_y-b_x p_z \\
\\
y^{\prime}=-\left(\frac{b_x b_y}{1+b_z}\right) p_x+\left(1-\frac{b_y^2}{1+b_z}\right) p_y-b_y p_z
\end{array}\right.
\label{xy_equ}
\end{equation}

Similarly, when $(x',y')$ and the direction vector $\hat{b}$ are given, the equation for any point $p$ on the $L$ is shown in equation \ref{p_equ}.

\begin{equation}
p=\left(\begin{array}{l}
1-b_x^2 /\left(1+b_z\right) \\
-b_x b_y /\left(1+b_z\right) \\
-b_x
\end{array}\right)\cdot x^{\prime}+\left(\begin{array}{l}
-b_x b_y /\left(1+b_z\right) \\
1-b_y^2 /\left(1+b_z\right) \\
-b_y
\end{array}\right)\cdot y^{\prime}
\label{p_equ}
\end{equation}

We can see from equation \ref{p_equ} that using Robert's representation, a line can be described with only 5 parameters after eliminating one redundant parameter.

Subsequently, for the computer to count votes, we must find a discrete parameter space. We begin by discretizing the direction vectors $\hat{b}$. Discretizing the three dimensions $b_x, b_y, b_z $ of $\hat{b}$ separately will result in a huge amount of computation. So we use the vertices coordinates of the polyhedra as the discrete direction vectors. More complex polyhedra lead to higher accuracy, but also to higher computational effort. It is necessary to choose the appropriate polyhedra according to the requirement. We use a polyhedron with 2562 vertices to improve the accuracy of trajectory extraction.

Subsequently, we discretize the plane $\beta$. Since the discretized direction vector is on a sphere centered at the origin, to find the range of $\beta$, it is necessary to translate the centers of 3D points $(x_i, y_i, z_i)$ to the origin. The range of intersection coordinates $x', y'$ between $\beta$ and the direction vector are $[-d/2,d/2]$. The formula for $d$ is :
\begin{equation}
d=\sqrt{(x_{max}-x_{min})^2+(y_{max}-y_{min})^2+(z_{max}-z_{min})^2}
\label{d_equ}
\end{equation}

Where $x_{max}, y_{max}, z_{max}, x_{min}, y_{min}, z_{min}$ denote the maximum and minimum of the 3D points $(x_i, y_i, z_i)$ on the x, y and z -axes, respectively.

\begin{figure}[h]
    \centering
    \includegraphics[width=2.5in]{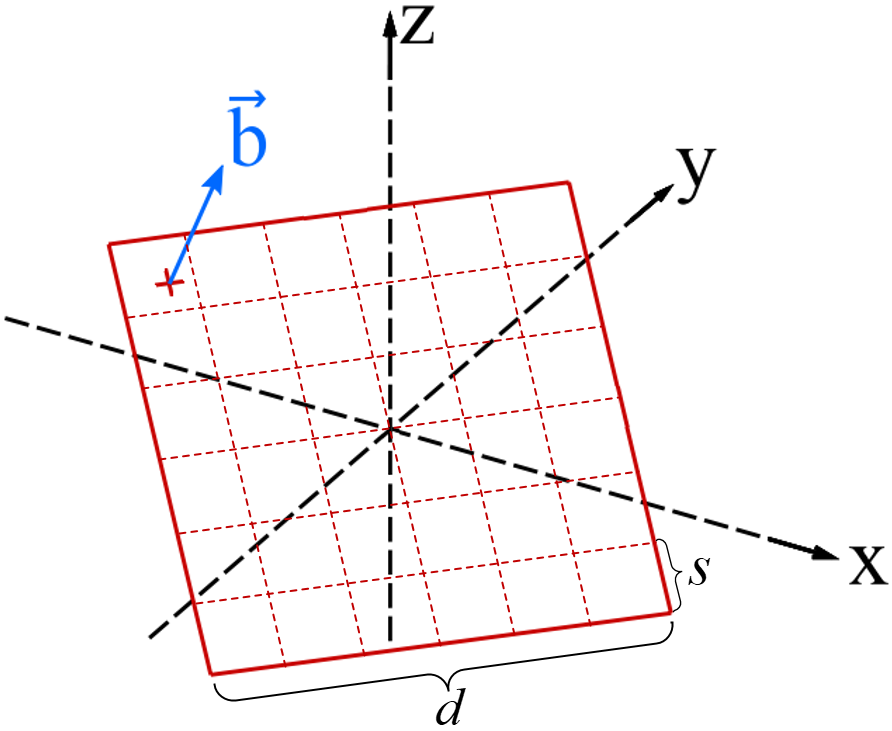}
    \caption{The discretization of plane.}
    \label{beta_plane_fig}
\end{figure}

Discretize the $\beta$ with a step size of $s$. Thus the voting space size of $\beta$ is $[d/s]*[d/s]$, as shown in Fig.\ref{beta_plane_fig}.

By using Robert representation, direction vector discretization, and plane $\beta$ discretization, the parameter space of the Hough transform is reduced to three dimensions, which greatly reduces the complexity. The number of voting grids in the parameter space is:
\begin{equation}
[d/s]*[d/s]*N_d
\label{girds_equ}
\end{equation}
where $N_d$ is the number of discretized directions.

The trajectory extraction method based on 3D Hough transform is shown in the algorithm \ref{3DHough_method}. After inputting a set of 3D points, we first translate the center of the 3D points to the origin. The parameter space is then discretized to create voting grids. Each point is transformed to the parameter space according to equation \ref{xy_equ}. Subsequently, votes are cast on the grid and the maximum votes will be found, if that maximum votes is greater than the voting threshold, a line is considered possible, otherwise there will be no line. After that, we get the anchor point in 3D space corresponding to the maximum votes and then calculate the equation of a candidate line in 3D space. We find the candidate points near the candidate line. If the number of candidate points is greater than the points threshold, a line is determined to exist and the next step is taken, otherwise, no line exists and the process is terminated. Finally, the candidate line is saved, the used candidate points and the votes are removed from parameter space. After all the lines are detected, we get the extraction results of the trajectory.

\begin{algorithm}
\caption{Trajectory Extraction Method Based on 3D Hough Transform}\label{3DHough_method}
\KwIn{3D points of trajectories: $P_{3D}$. Grid Size: $S_g$. Minimum number of votes in the voting space: $V_{min}$. Minimum number of candidate points for a line: $N_{P}$.}
\KwResult{Trajectory position matrix after trajectory extraction: $P_E$.}
$Lines=\emptyset$\\
$P_{3D}=TranslateCenterToOrgin(P_{3D})$\\
$\beta, \hat{b}=CreateParameterSpace(S_g)$\\
$P'_{3D}=TransformToParameterSpace(P_{3D})$\\
\While{True}{
$V_{max}, \vec{V}, P = GetMaxVotesLine()$\\
\If{$V_{max}<V_{min}$}{
break
}
$P_C,P_{CI} = PointsCloseToLine(\vec{V}, P)$\\
\If{$Len(P_C)<N_{P}$}{
break
}
$\vec{V}, A = orthogonalLSQ(P_C)$\\
$P_C,P_{CI} = PointsCloseToLine(\vec{V}, A)$\\
\If{$Len(P_C)<N_{P}$}{
break
}

$RemovePointsFromParameterSpace(P_{CI})$\\
$Lines=Lines\cup P_{CI}$
}
$P_E=0$\\
\For{$i=1:Len(Lines)$}{
$Line=Lines[i]$\\
$P_E[Line]=255$
}
return $P_E$

\end{algorithm}

\subsection{Trajectory-based Tracking Method for Multiple Small Targets}
\begin{figure}[h]
    \centering
    \includegraphics[width=3.5in]{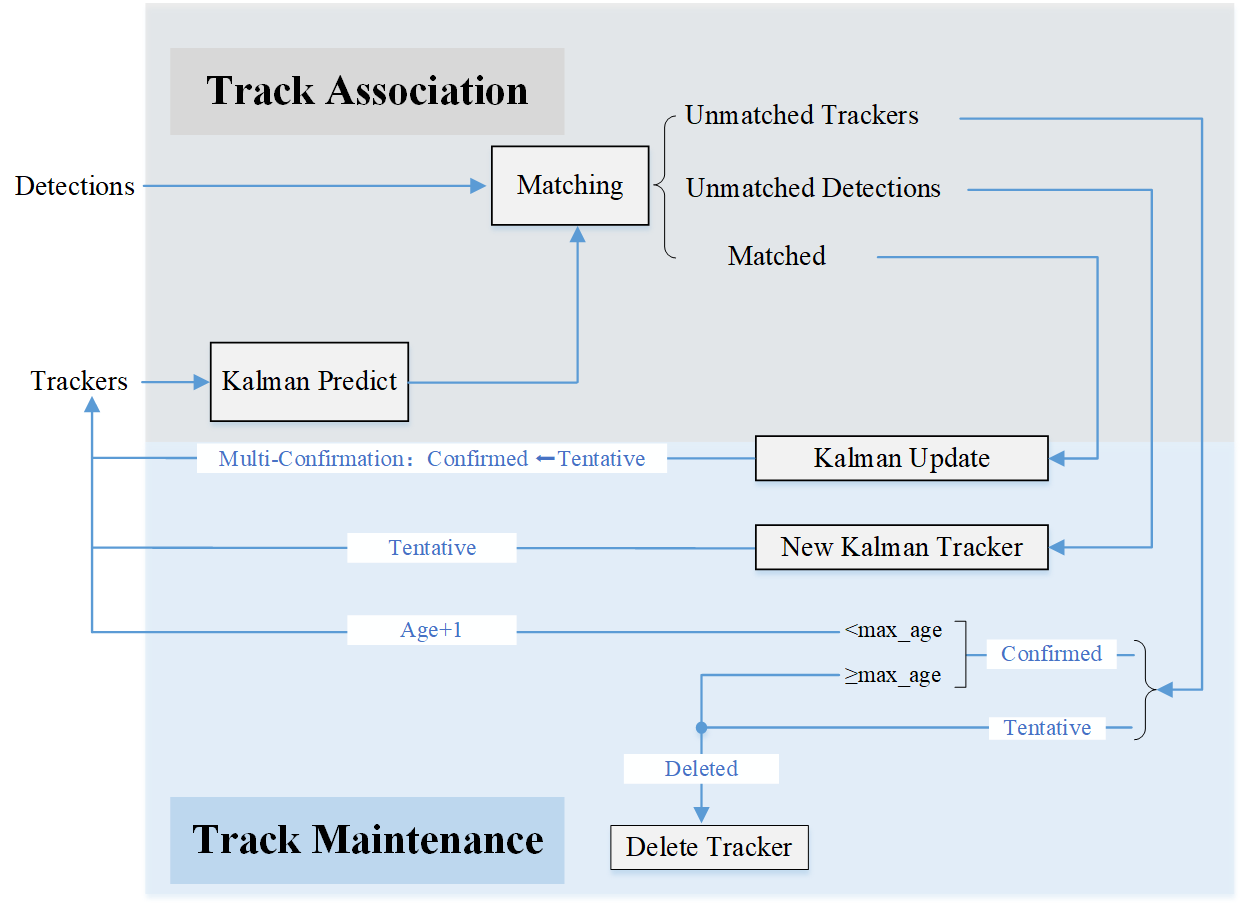}
    \caption{The workflow of our trajectory-based tracking method.}
    \label{TrackProcess_fig}
\end{figure}

Tracking the small and weak target in a single frame is a very difficult task due to the lack of spatial features. Instead of tracking small targets directly, we track target trajectories of multiple frames. This greatly reduces the difficulty of tracking due to the uniqueness of the target trajectory and the richness of its spatial features. The problem with tracking trajectories is that false alarm points in the detection results affect the accuracy of the tracking. However, we have eliminated the false alarm points by trajectory extraction, which provides the conditions for our trajectory-based multiple small targets tracking method. We were inspired by the tracking strategy of DeepSORT\cite{Wojke_2017_Simple} to design our tracking method, as shown in Fig.\ref{TrackProcess_fig}.

Our tracking method consists of Kalman filtering and trajectory matching. We use Kalman filtering to predict and update the tracking results, and the cost-matching algorithm is used to associate the trackers with the detection trajectories. We introduce a state description of the trajectory and a multiple confirmation strategy to cope with mistracking and disappearance of the target. We first introduce the trajectory-based Kalman filter, followed by the cost-matching algorithm, and finally our tracking strategy.
\subsubsection{Trajectory-based Kalman Filter}
Kalman filtering is a classical tracking algorithm, which consists of two parts: predict and update. Kalman filtering first predicts the next state of the target based on the model, and then updates the predicted state when the next detection result is obtained to get a more accurate result. Correctly describing the state of the target is a very important task. We use the center position $(P_x, P_y)$ and velocity $(V_x,V_y)$ as the state of the target trajectory. It is worth mentioning that since the number of frames in our detection unit is constant, the velocity of the target with fixed time is the length of the target trajectory in a given direction. Using target center position and velocity as states is enough, and it is unnecessary to add target appearance states. We give the prediction and update equations for Kalman filtering.
\begin{equation}
\left\{\begin{array}{l}
\hat{x}_k=F x_{k-1}+w \\
\hat{P}_k=F P_{k-1} F^T+Q
\end{array}\right.
\label{kalman_predict_equ}
\end{equation}

In the above equation, $x_{k-1}$ denotes the state of the target at the time of $k-1$; $F$ denotes the state transfer matrix; $w$ denotes the noise of motion model; $\hat{x}_k$ denotes the predicted state of the target at the time of k; $P_{k-1}$ denotes the covariance of the states at $k-1$ time; $\hat{P}_k$ denotes the covariance of the predicted states at $k$ time. The difference $y_k$ between the observation $z_k$ (target detection result) and the prediction at $k$ time is:
\begin{equation}
y_k=z_k-H\hat{x}_k
\label{yk_equ}
\end{equation}

Where $H$ is used to transform the state space to the observation space. The Kalman gain $K$ at $k$ time is:
\begin{equation}
K=\hat{P}_kH^T(H\hat{P}_kH^T+R)^{-1}
\label{kalman_gain_equ}
\end{equation}

$R$ denotes the measurement noise, i.e., the noise of the target detection process. Based on the Kalman gain, we can get the equation of the updating process:
\begin{equation}
\left\{\begin{array}{l}
x_k=\hat{x}_k+Ky_k \\
P_k=(I-KH)\hat{P}_k
\end{array}\right.
\label{kalman_update_equ}
\end{equation}

The purpose of the update process is to correct the prediction based on the observations (results of the target detection) to get a more accurate estimate of the state, which is $x_k$; $P_k$ denotes the covariance of the final estimated state.

Now that we have constructed the Kalman filter, we can use it for tracking.
\subsubsection{Cost Matching Algorithm}
When tracking multiple targets, each target has its own Kalman tracker. Suppose there are currently $M$ trackers but $N$ targets are detected. How to rightly associate M trackers with N targets is an important issue. The target association problem during multi-target tracking is usually modeled as a minimum cost estimation of a two-dimensional matching problem. Design cost matching formulas are important.

We use $C_{i,j}$ to denote the matching cost between the target $T_i$ detected at time $k$ and the tracker $K_j$ at time $k-1$. The smaller the $C_{i,j}$, the higher the probability of tracking correctly. Hungarian algorithm is used to match the detection results with the tracker at minimum cost. We give the calculation of the cost matrix, as shown in equation \ref{cost_equ}.
\begin{equation}
\left\{\begin{array}{l}
c_1=\sqrt{(x_k-\hat{x}_{k})^2+(y_k-\hat{y}_{k})^2} \\
c_2=1-\frac{\hat{v}_{k}\cdot v_k}{||\hat{v}_{k}|| ||v_k||}, v=(v_x,v_y) \\
c=\alpha_1 *c_1+\alpha_2 *c_2
\end{array}\right.
\label{cost_equ}
\end{equation}

Where $c_1$ is the cost between the predicted position $(\hat{x}_{k},\hat{y}_{k})$ and the detected position $(x_k,y_k)$, calculated using the Euclidean distance. $c_2$ is the cost between the predicted velocity $\hat{v}$ and the detected velocity $v$, calculated using the cosine distance since the velocity is a vector. $c$ is the total cost, which is the weighted sum of $c_1$ and $c_2$. The cost matching matrix $C$ is shown as equation \ref{cost_matrix_equ}, $c_{i,j}$ is the cost between tracker $i$ and detection result $j$. We use the minimum $c_{i,j}$ to match the tracker with the detection result. To prevent tracking errors, we set the cost threshold. If the minimum cost is too large, a new tracker will be created.

\begin{equation}
C=\left[\begin{array}{l}
c_{1,1}, \ldots \ldots ., c_{1, N} \\
\ldots \ldots, c_{i, j}, \ldots \ldots \\
c_{M, 1}, \ldots \ldots, c_{M, N}
\end{array}\right]
\label{cost_matrix_equ}
\end{equation}
\subsubsection{Tracking Strategy}
Setting up a sensible tracking strategy is critical to tracking targets correctly. We introduce three states to address the question of whether the tracker is retained when the target disappears. The three states are Tentative, Confirmed, and Deleted. Next, we detail our tracking strategy as shown in Fig \ref{TrackProcess_fig}.

When a new target is detected, a tracker will be created to track this target, which has a tentative state. Subsequently, this tracker predicts the state of the tracked target at the next moment and matches it with the next detection state, and if the multiple matches are right, the state of this tracker is switched to confirmed. The difference between the tentative state and the confirmed state is the deletion process of the tracker. If the state of a tracker is tentative, the tracker will be deleted when the tracker loses the target once. While the state of the tracker is confirmed, the tracker will be deleted until the number of times the tracker loses the target reaches $max_age$. This strategy ensures that the tracker is not deleted if the target briefly disappears.

\section{Experimental Analysis}
\begin{table*}[t]
\centering
\caption{Detail of Sequences}
\label{seq_detail}
\begin{tabular}{cccccccc}
  \hline
  Sequence & Frames & Image Size & Average SCR & Target Size & Target Description & Background Description    \\
  \hline
  1   & 794 & $128\times128$ & 0.69 & $9\sim 12$ pixels & Single mini-car, 0.12 pixels/frame & Ground, trees\\
  2   & 420 & $256\times256$ & 0.34 & $6\sim 12$ pixels & Single UAV, 0.21 pixels/frame & Mountains, buildings, trees\\
  3   & 683 & $256\times256$ & 0.37 & $4\sim 9$ pixels & Single simulation target, 0.23 pixels/frame & Clouds\\
  4   & 1000 & $256\times256$ & 1.25 & $11\sim 32$ pixels& Single aircraft, 0.26 pixels/frame & Power poles, buildings\\
  5   & 1024 & $256\times256$ & 1.61 & $18\sim 50$ pixels  & Two aircrafts, 0.13/0.19 pixels/frame & Wires, buildings\\
  6   & 700 & $128\times128$ & 0.42 & $15\sim 36$ pixels  & Single asteroids, 0.15 pixels/frame & Stars\\
  \hline
\end{tabular}
\end{table*}

\begin{table*}[htpb]
\centering
\caption{Detailed Parameter Settings of Comparison Methods }
\label{methods_detail}
\begin{tabular}{cl}
  \hline
  Method & Parameters    \\
  \hline
  Top-Hat\cite{Zeng_2006_design} & step=40, min size=1, max size=3 \\
  LCM\cite{Chen_2014_Local} & structure size=3 \\
  MPCM\cite{Wei_2016_Multiscalea} & max size=3 \\
  IPI\cite{Gao_2013_Infrared} & patch size $m_1 \times n_1: 20 \times 20$, sliding step: 10, $\lambda=1/\sqrt{\text{min}(m_1,n_1)}$, $\varepsilon=1e-7$ \\
  PSTNN\cite{Zhang_2019_Infrared} & patch size $m_1 \times n_1: 20 \times 20$, sliding step: 10, $\lambda=1/\sqrt{\text{max}(n_1,n_2)*n_3}$, $\varepsilon=1e-7$ \\
  NRAM\cite{Zhang_2018_Infrared} & patch size $m_1 \times n_1: 20 \times 20$, sliding step: 10, $\lambda=1/\sqrt{\text{min}(m_1,n_1)}$, $\varepsilon=1e-7$ \\
  NAF\cite{Liu_2015_Moving} & Seq1, Seq 4, Seq 5: window size=61; Seq 2:window size=31; Seq 3: window size=41; Seq 6: window size=71\\
  ICLSP\cite{Liu_2015_Temporal} & $\sigma = 2, g_c=7, \sigma_{\tau}=200$\\
  TRLCM\cite{Han_2022_Small} & P=20, G=10, B=10\\
  KF\cite{Wu_2018_Weak} & Seq1, Seq 4, Seq 5: window size=60; Seq 2:window size=30; Seq 3: window size=40; Seq 6: window size=70\\
  Ours TESS & Seq1, Seq 4, Seq 5: window size=60; Seq 2:window size=30; Seq 3: window size=40; Seq 6: window size=70\\
  \hline
\end{tabular}
\end{table*}

\begin{figure*}[h]
\centering
\includegraphics[width=6.8in]{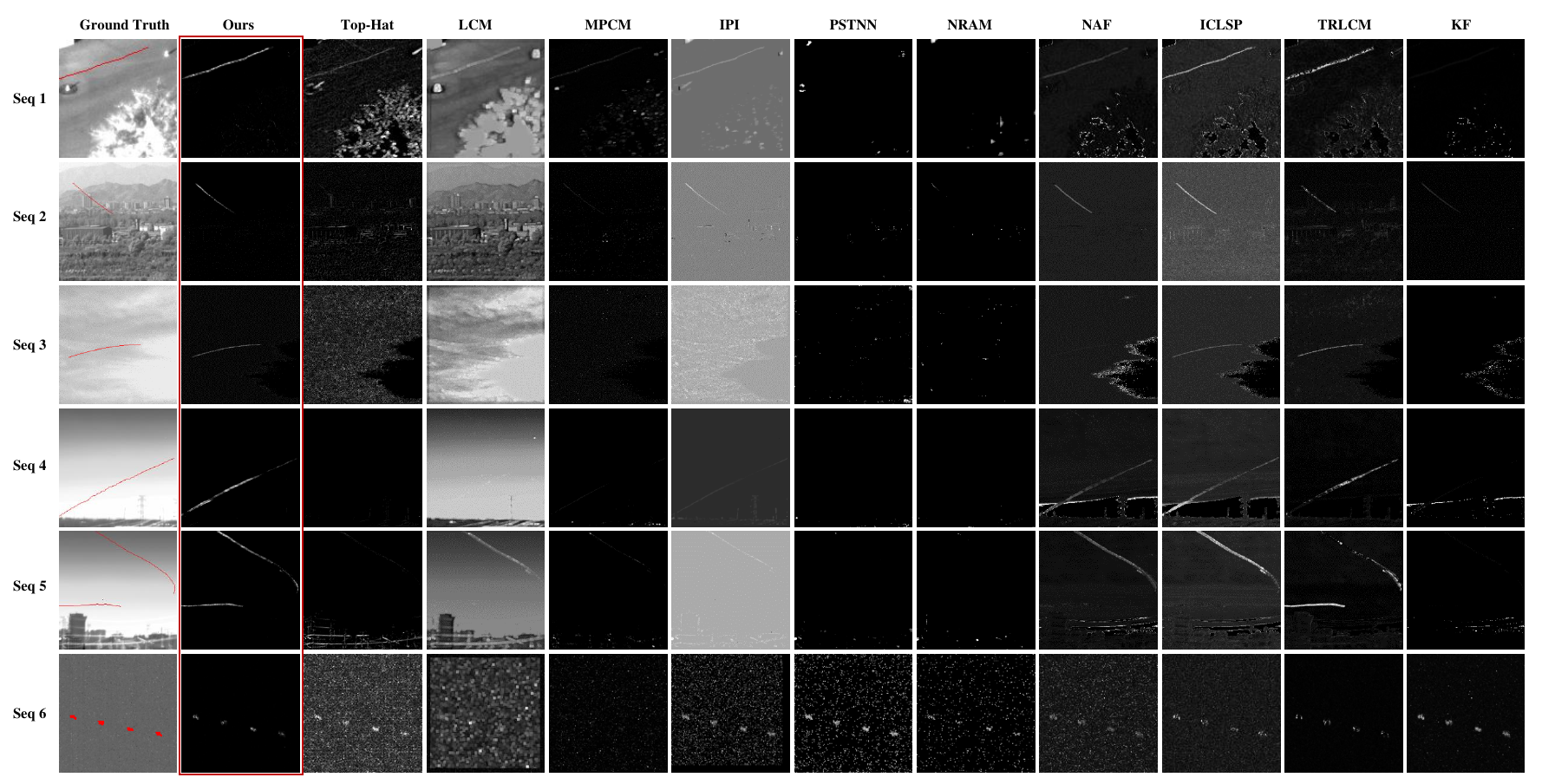}
\caption{Visual results of all comparison methods on sequence 1-6.}
\label{visual_results_fig}
\end{figure*}

\begin{figure*}[t]
\centering
\includegraphics[width=7in]{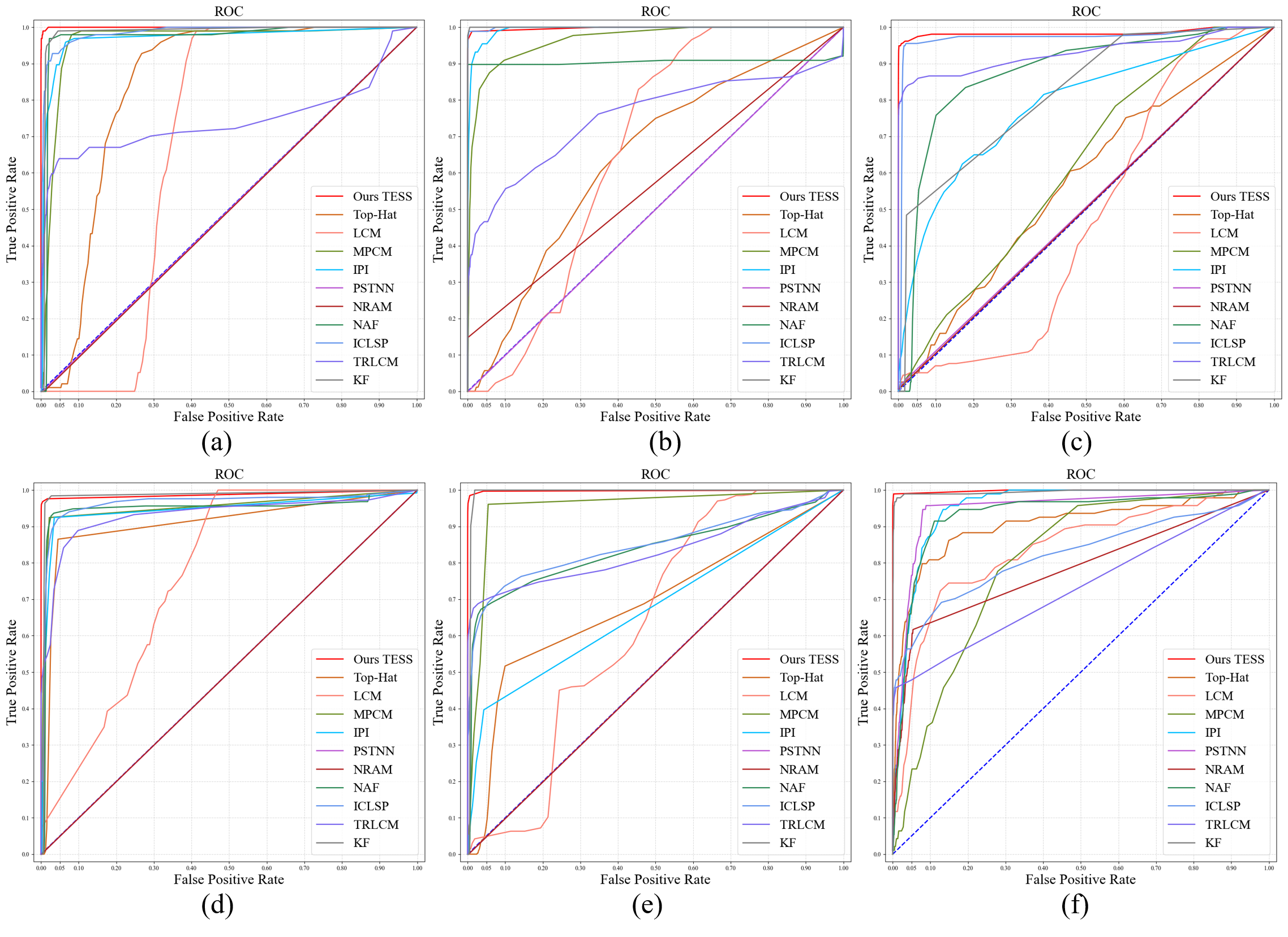}
\caption{ROC curves of the six sequences. (a)–(f) Curves correspond to the sequences 1–6 in Table \ref{seq_detail}. When the curve is closer to the upper left corner or has greater AUC, it means that the method has better detection performance.}
\label{ROC_fig}
\end{figure*}

\begin{table*}[h]
\centering
\caption{Comparison of Different Methods for Quantification of Sequences 1-6}
\label{method_metrics}
\begin{tabular}{c|c|c|c|c|c|c|c|c|c|c|c|c}
  \hline
  Metrics  & \multicolumn{6}{c|}{SCRG} & \multicolumn{6}{c}{BSF}\\
  \hline
  Sequence&1&2&3&4&5&6&1&2&3&4&5&6\\
  \hline
  Ours TESS&\textbf{65.22}&\textbf{241.49}&\textbf{42.65}&\textbf{101.62}&\textbf{115.34}&  \textbf{220.32}&\textbf{13.06}&\textbf{68.42}&\textbf{3.17}&\textbf{80.12}&\textbf{101.49}&\textbf{8.05} \\
  Top-Hat\cite{Zeng_2006_design}&1.15&0.88&0.68&1.05&0.09& 6.7&0.99 &3.11&0.52& 25.46&2.93&0.15\\
  LCM\cite{Chen_2014_Local}& 0.52&1.43&0.11&0.7     &0.3&3.99&1.09 &1.29&0.51&0.94&1.7&0.17\\
  MPCM\cite{Wei_2016_Multiscalea}&3.56& 14.49&0.71&1.54& 0.93&2.1&3.7& 9.25& 1.67&10.69& 9.21&0.4\\
  IPI\cite{Gao_2013_Infrared}&6.07&27.57&3.79&3.47 &0.78&  6.9&6.11& 9.67&1.24&24.24& 15.8&0.16\\
  PSTNN\cite{Zhang_2019_Infrared}&0.09&0.11& 0.04&0.03 &0.02&7.98&3.03& 8.89&1.31&7.87& 7.21&0.13\\
  NRAM\cite{Zhang_2018_Infrared}&0.12&17.66&0.48&0.02 &0.03&6.27&3.44&12.41&2.19&11.23& 7.08&0.18\\
  NAF\cite{Liu_2015_Moving}&2.51&46.28&0.84&1.65 &1.34&6.47&2.01&10.93&0.7&2.5&3.25&0.25\\
  ICLSP\cite{Liu_2015_Temporal}&8.59&26.41&9.31&3.34 &3.09&12.0&2.03&3.41&1.01&3.14&3.97&0.46\\
  TRLCM\cite{Han_2022_Small}&7.05&12.78&26.07&9.12 &9.0&44.42&2.67&5.07&1.71&10.26& 7.32&2.28\\
  KF\cite{Wu_2018_Weak}      &1.02&65.49&0.02&0.25  &0.22&37.63&3.16&31.66&0.96&2.9&4.4&0.82\\
  \hline
\end{tabular}

\vspace{2mm} 

\begin{tabular}{c|c|c|c|c|c|c|c|c|c|c|c|c}
  \hline
  Metrics  & \multicolumn{6}{c|}{AUC}& \multicolumn{6}{c}{Running Time (frame/ms) } \\
  \hline
  Sequence&1&2&3&4&5&6&1&2&3&4&5&6    \\
  \hline
   Ours TESS    
  &\textbf{0.9997}  &0.9978 &\textbf{0.9843} &\textbf{0.9877} &\textbf{0.9981} & \textbf{0.9983}
  &1.21&2.26&\textbf{1.48}&\textbf{1.33}&\textbf{1.32}&1.29\\
  Top-Hat\cite{Zeng_2006_design}
  &0.8359  &0.6431 &0.5747 &0.9080 &0.6884 & 0.9053
  &\textbf{0.43}&\textbf{1.93}&2.46&2.06&2.78&\textbf{0.37}\\
  LCM\cite{Chen_2014_Local}     
  &0.6766  &0.6636 &0.4752 &0.7647 &0.6309 & 0.8363
  &1109.50&4827.02&4792.96&4805.97&4798.63&1114.03\\
  MPCM\cite{Wei_2016_Multiscalea}
  &0.9615  &0.9662 &0.6197 &0.9541 &0.9526 & 0.8028
  &1317.58&5512.67&5462.09&5402.68&5519.20&1289.46\\
  IPI\cite{Gao_2013_Infrared}
  &0.9695  &0.9937 &0.7836 &0.9454 &0.6780 & 0.9529
  &1398.83	&6870.74	&6616.40	&20660.57&	22369.25&	843.04\\
  PSTNN\cite{Zhang_2019_Infrared}
  &0.4982  &0.4992 &0.5059 &0.4993 &0.4988 & 0.9474
  &49.31 &	373.98 &	334.90& 	233.96& 	605.47& 	59.80 \\
  NRAM\cite{Zhang_2018_Infrared}
  &0.4957  &0.5736 &0.5026 &0.4993 &0.4987 & 0.7850
  &86.59 &	546.74 &	351.27 &	557.09& 	562.35 &	94.84 \\
  NAF\cite{Liu_2015_Moving}     
  &0.9717  &0.9050 &0.8749 &0.9491 &0.8376 & 0.9311
  &612.58 &	2402.62 	&2447.72 	&2472.54 &	2509.79 &	600.30 \\
  ICLSP\cite{Liu_2015_Temporal}
  &0.9860  &0.9992 &0.9708 &0.9647 &0.8494 & 0.8233
  &13.21& 	58.05 &	53.13 	&51.45 &	52.32 &	13.73 \\
  TRLCM\cite{Han_2022_Small}
  &0.7454  &0.7556 &0.9302 &0.9322 &0.8333 & 0.7324
  &228.93 &	851.79 &	898.62 &	921.51 &	940.46 &	225.77 \\
  KF\cite{Wu_2018_Weak}      
  & 0.9859 &\textbf{0.9999} &0.8210 &0.9839 &0.9928 & 0.9953
  &15.26 &	28.71 &	38.92 	&55.60 &	60.18 &	17.50 \\
  \hline
\end{tabular}
\end{table*}

In this section, we evaluate the performance of the proposed detection and tracking methods through a series of experiments. Firstly, we test the proposed detection method based on temporal energy selective scaling and compare it with other detection methods to demonstrate its superiority. Afterward, we validate the ability of the proposed trajectory extraction method to reduce false alarms. Finally, we verify the capability of the multi-target tracking method.
\subsection{Experiments of Detection Method}
In this section, we evaluate the performance of the proposed detection method based on temporal energy selective scaling (TESS). We first give the evaluation metrics. Next, we describe in detail the sequences used for the experiment. Finally, we perform the comparison experiment with other classical methods.

\subsubsection{Evaluation Metrics}
In this article, we use three commonly used metrics to evaluate the performance of small target detection methods, including the receiver operating characteristic (ROC) curve, the signal-to-clutter ratio gain (SCRG), and the background suppression factor (BSF).

\begin{itemize}
    \item Receiver Operating Characteristic (ROC): The ROC curve is a widely used assessment tool in target detection tasks. In this curve, the vertical axis represents the true positive rate (TPR), while the horizontal axis denotes the false positive rate (FRP). These two parameters are defined as follows:

\begin{align}
\text{FPR}=\frac{\text{Pixels of False Targets Detected}}   {\text{Pixels of Actual Background}} \nonumber \\
\text{TPR}=\frac{\text{Pixels of Real Targets Detected}}{\text{Pixels of Actual Targets}}
\label{TPR_FPR_equ}
\end{align}
    
    In general, the closer the ROC curve is to the upper left corner or the larger the area under the curve (AUC), the better the detection capability of the method.
    \item Signal-to-Clutter Ratio Gain (SCRG): The SCRG is calculated by the formula:
    \begin{equation}
    \text{SCRG}=\frac{\text{SCR}_\text{out}}{\text{SCR}_\text{in}}
    \label{SCRG_equ}
    \end{equation}
    where $\text{SCR}_\text{in}$ is the average SCR of the original image sequence, $\text{SCR}_\text{out}$ is the average SCR of detection result. The SCRG indicates the ability of the method to enhance the target and suppress the background.
    \item Background Suppression Factor (BSF): The BSF is a metric used to express the ability of methods to suppress background, as shown below:
    \begin{equation}
    \text{BSF}=\frac{\sigma_{\text{in}}}{\sigma_{\text{out}}}
    \label{BSF_equ}
    \end{equation}
    where $\sigma_{\text{in}}$ and $\sigma_{\text{out}}$ are the average background standard deviation of the original image sequence and the detection result.
\end{itemize}

Higher SCRG and BSF indicate better performance of the method.

\subsubsection{Sequences for Experiment}
We used six image sequences for the experiment and the detail of the six sequences is shown in Table \ref{seq_detail}. Sequence 1 is the movement of a mini-car on the ground captured by VLC (Visible Light Camera), with a tree present in the scene as clutter. Sequence 2 is the motion process of a small UAV captured by VLC in a complex background with mountains, buildings, and trees. Sequence 3 is the motion process of a target in clouds. Since targets in clouds are difficult to capture, a Gaussian target generated according to the PSF, as shown in Fig. \ref{target_fig} (a), is added to the cloud background captured by VLC frame by frame. Sequence 4 and Sequence 5 show the flight process of aircraft above an airport captured by a mid-wave infrared camera. Sequence 4 contains one aircraft and Sequence 5 contains two aircrafts. Sequence 6 is an asteroid motion process captured at an observatory using a telescope. 

These sequences contain visible data, infrared data, and astronomical data. The image size, frames, SCR, target size, and target velocity of sequences are all different. All six sequences contain interferences in the background. So using these six sequences is enough to evaluate the robustness of our method in different scenes and targets.

\subsubsection{Baseline Methods}
To demonstrate the excellent performance of our method, which is compared with ten other small target detection methods in the above sequences, including Top-Hat\cite{Zeng_2006_design}, LCM\cite{Chen_2014_Local}, MPCM\cite{Wei_2016_Multiscalea}, IPI\cite{Gao_2013_Infrared}, PSTNN\cite{Zhang_2019_Infrared}, NRAM\cite{Zhang_2018_Infrared}, NAF\cite{Liu_2015_Moving}, ICLSP\cite{Liu_2015_Temporal},TRLCM\cite{Han_2022_Small}, KF\cite{Wu_2018_Weak}. Among them, Top-Hat is a filter-based detection method, while LCM and MPCM are HVS-based methods. IPI, PSTNN, and NRAM are LRSD-based methods. NAF, ICLSP, TRLCM, and KF are ITP-based methods. Table \ref{methods_detail} lists the detailed parameter settings of the comparison method. The parameters of the detection method are set according to the spatio-temporal features of the target and are tuned around empirical values to obtain the best results. In the single-frame detection method, the parameters are set based on the size of the target. In the ITP-based method, the parameter setting is based on the length of the target signal, which is determined by the speed of the target and the frame rate of the camera. When the target speed is unknown, one-tenth of the length of the image sequence can be used as the window size first and subsequently adjusted according to the detection results. PSTNN and NRAM were implemented by Matlab, others were implemented by Python. The computer used for the experiment has a CPU of Intel Core i7-10700K and 32 GB RAM. The graphics card used in the method of parallelization acceleration is the NVIDIA GeForce RTX 3080.

\subsubsection{Visual Results}

Figure \ref{visual_results_fig} shows the visualization of the detection results for all compared methods. All the detection results are unthresholded, which can better reflect the ability of the detection methods to suppress the background and enhance the target. Top-Hat detected the moving targets in sequence 1 and sequence 6, but there was a lot of background clutter in the detection results, and the targets in sequences 2-5 were not detected at all. LCM only detected the moving targets in sequence 1 and sequence 5. Since the detection results are not thresholded, LCM only calculates the contrast of the image, and a large amount of background is still retained in the detection results. As an improved method of LCM, the background is suppressed a lot in the detection result of MPCM, but it only detects the target in sequence 1, the target trajectory is still weak, and the rest of the targets are suppressed as background. In the LRSD-based methods, IPI had the best detection performance, with targets detected in all sequences except sequence 3. However, a lot of background clutter was still retained in the detection results. PSTNN and NRAM only detected targets in sequence 6, and the rest of the targets were suppressed as background. The ITP-based detection method has better detection performance. In the NAF detection results, targets were detected in all sequences except sequence 3. ICLSP detected most of the targets, but not one of the targets in sequence 5. TRLCM detected most of the targets. But there is too much background clutter in NAF, ICLSP, and TRLCM. And KF did detect the target, but the trajectory was too weak.

The performance of the detection method is related to the size and SCR of the target. Most of the methods detected the target in sequence 1, which is due to the larger size and higher SCR of the target. The SCR of sequence 2 and sequence 3 are close because the target in sequence 2 is bigger, most methods detect sequence 2 better than sequence 3. The detection results of most methods in sequences 5 and 6 are similar, but except for our method and TRLCM, other methods did not detect the horizontally moving target in sequence 5, which is due to the SCR of this target being too low. Sequence 6 has a larger target size and a cleaner background, so most of the methods detected the target.

Our method detected all targets with a well-suppressed background, which demonstrates the excellent performance of our method for small targets with low SCR in different scenarios. The intensity of the target trajectory is not always the same because the SCR changes dynamically as the target moves. Since the background is cleanly suppressed, the target and background can be separated using simple threshold segmentation.

\begin{figure*}[t]
\centering
\includegraphics[width=7in]{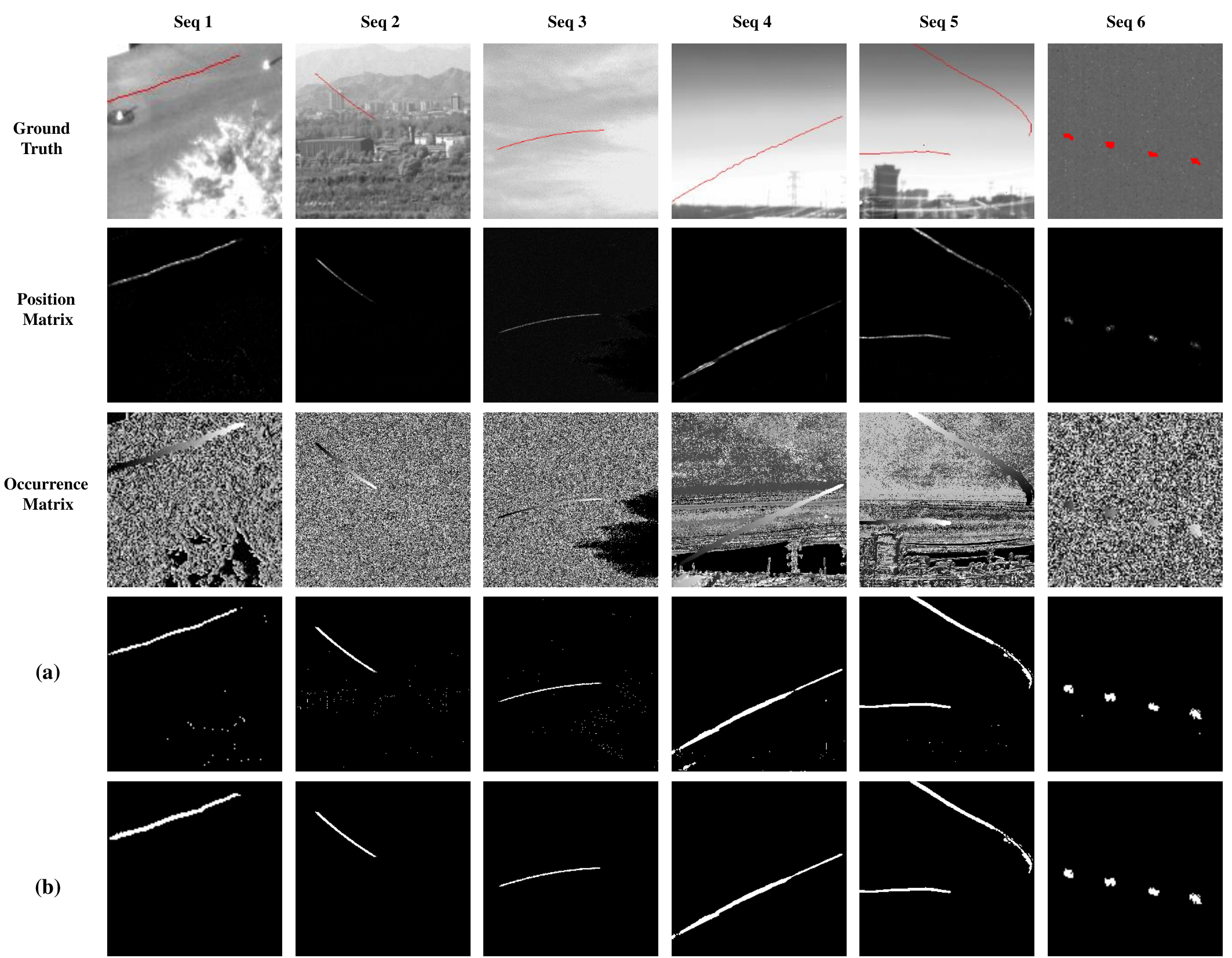}
\caption{Visualization results for target detection and trajectory extraction. (a) Traditional thresholding results. (b) Results using our trajectory extraction method.}
\label{Track_Extraction_fig}
\end{figure*}

\begin{table}[t]
\centering
\caption{Detailed Parameter Settings of Trajectory Extraction Method.}
\label{Track_Extract_methods_detail}
\begin{tabular}{cl}
  \hline
  Sequence & Parameters\\
  \hline
  1 & $\theta_I=4, S_g=4,V_{min}=20, V_P=20$\\
  2 & $\theta_I=4, S_g=8,V_{min}=10, V_P=10$\\
  3 & $\theta_I=10, S_g=4,V_{min}=20, V_P=20$\\
  4 & $\theta_I=4, S_g=20,V_{min}=50, V_P=200$\\
  5 & $\theta_I=4, S_g=20,V_{min}=20, V_P=20$\\
  6 & $\theta_I=4, S_g=20,V_{min}=5, V_P=5$\\
 \hline
\end{tabular}
\end{table}

\begin{table*}[t]
\centering
\caption{Metrics of traditional thresholding method and our trajectory extraction method.}
\label{Track_Extract_method_metrics}
\begin{tabular}{c|c|c|c|c|c|c|c|c|c|c|c|c}
  \hline
    & \multicolumn{12}{c}{Sequence}\\
  \hline
    & \multicolumn{2}{c|}{1}& \multicolumn{2}{c|}{2}& \multicolumn{2}{c|}{3}& \multicolumn{2}{c|}{4}& \multicolumn{2}{c|}{5}& \multicolumn{2}{c}{6}\\
\hline
  Method &TPR&FPR&TPR&FPR&TPR&FPR&TPR&FPR&TPR&FPR&TPR&FPR\\
  \hline
  Ours TE & \textbf{96.91\%}&\textbf{0.00\%}&\textbf{98.86\%}&\textbf{0.00\%}&\textbf{94.27\%}&\textbf{0.00\%}&\textbf{96.83\%}&\textbf{0.00\%}&\textbf{94.29\%}&\textbf{0.00\%}&\textbf{97.87\%}&\textbf{0.00\%}\\
  Thresholding& 96.91\%&0.17\%&97.73\%&0.14\%&94.27\%&0.13\%&96.42\%&0.17\%&94.29\%&0.03\%&97.87\%&0.01\%
\\
    \hline
    Running Time of TE (ms)& \multicolumn{2}{c|}{168.58}& \multicolumn{2}{c|}{44.88}& \multicolumn{2}{c|}{223.40}& \multicolumn{2}{c|}{109.73}& \multicolumn{2}{c|}{123.64}& \multicolumn{2}{c}{11.94}\\
  \hline
\end{tabular}
\end{table*}

\begin{table*}[h]
\centering
\caption{Detail of Tracking Sequence.}
\label{Detail_Track_Seq_Table}
\begin{tabular}{ccccc}
  \hline
  Target & SCR & Size & Speed & Moving Description\\
  \hline
  1 &1.34 & $13\sim 20$ pixels  & 0.25 pixels/frame & \multicolumn{1}{m{11cm}}{ Take off from the airport at frame 400, move to the upper right, turn left at frame 1080, leave the field of view at frame 2300.} \\
  2 &2.43 &$16\sim 25$ & 0.16 pixels/frame & \multicolumn{1}{m{11cm}}{Take off at frame 600, fly right and up. Leave the field of view at frame 1540.} \\
  3 &1.35 &$8\sim 12$ pixels & 0.17 pixels/frame & \multicolumn{1}{m{11cm}}{Appears at frame 1060, moving from right to left, disappears at frame 4680.} \\
  4 &1.13 &$6\sim 9$ pixels & 0.19 pixels/frame & \multicolumn{1}{m{11cm}}{Appears at frame 3400, moves horizontally from right to left, disappears at frame 4480.} \\
  5 &0.83 & $9\sim 12$ pixels & 0.46 pixels/frame & \multicolumn{1}{m{11cm}}{Appears at frame 3440, moving horizontally from left to right, leaving the field of view at frame 4840.} \\ 
  6 &0.72 &$14\sim 26$ &0.30 pixels/frame & \multicolumn{1}{m{11cm}}{Take off at frame 3720, fly to the upper right, and leave the field of view at frame 4520.} \\
  \hline
\end{tabular}
\end{table*}

\begin{table*}[]
\centering
\caption{Detailed Parameter Settings of Tracking Methods }
\label{tracking_methods_detail}
\begin{tabular}{cl}
  \hline
  Method & Parameters    \\
  \hline
  Top-Hat\cite{Zeng_2006_design} & step=40, min size=1, max size=3 \\
  IPI\cite{Gao_2013_Infrared} & patch size $m_1 \times n_1: 50 \times 50$, sliding step: 20, $\lambda=1/\sqrt{\text{min}(m_1,n_1)}$, $\varepsilon=1e-7$ \\
  KCF\cite{Henriques_2015_High-Speed}& $h=w=10$\\
  CSRT\cite{Lukezic_2017_Discriminative}& $h=w=10$\\
  MIL\cite{Babenko_2011_Robust}& $h=w=10$\\
  BOOSTING\cite{Oza_2001_Online}& $h=w=10$\\
  MOOSE\cite{Bolme_2010_Visual}& $h=w=10$\\
  Ours&Unit length=500, window size=50, $\theta_I=18, S_g=12,V_{min}=30, V_P=40, \text{Age}_{\text{max}}=3, \text{Hit}_{\text{min}}=3$\\
  
  \hline
\end{tabular}
\end{table*}

\subsubsection{Quantitative Results}
To quantify the detection performance of different methods, we evaluated them using ROC, AUC, SCRG, and BSF as described above, the ROC curves are shown in Fig.\ref{ROC_fig} and the metrics are shown in Table \ref{method_metrics}. It can be seen that our method has the highest SCRG and BSF in all the sequences, which shows that our method can enhance the target and suppress the background well. Among the single-frame detection methods, IPI has the best detection performance. The detection performance of ITP-based methods is better than the single-frame detection method. This is consistent with the results of Fig.\ref{visual_results_fig}. In the ROC curves, our method has the largest AUC except for sequence 2, which indicates that in most of the sequences, our method can achieve the highest detection rate at the lowest false alarm rate. Although the AUC of the ICLSP and KF methods in Sequence 2 are larger than ours, the ability of ICLSP to suppress the background and the ability of KF to enhance the target are both poorer. Thus, taken together our method has the best detection performance.

The speed of detection is also very important. We have counted the average running time of different methods on a single frame. Among them, our method is fastest on sequences 3, 4, and 5, and slower than the Top-Hat on the other sequences. Top-Hat is essentially a filter and the open and close operations in morphology are accelerated by the OpenCV library, so it has a faster detection speed. Moreover, Top-Hat's detection result is almost unusable. The detection speed of other methods is very slow, and some methods detect a frame even more than 5 seconds, which cannot meet the actual demand.

Therefore, our detection method achieves the best detection results with very low detection time consumption and has great potential for application.

\subsection{Experiments of Trajectory Extraction Method}

We conduct experiments about the trajectory extraction (TE) method based on the detection results of sequences 1-6 to verify the advantages of reducing the false alarm rate. As shown in Algorithm \ref{detection_method}, the detection result of our method consists of a position matrix and an occurrence time matrix, as shown in Fig.\ref{3D_trace} (a) and (b). The two matrices are fused for trajectory extraction. The parameters used for trajectory extraction in sequences 1-6 are shown in Table \ref{Track_Extract_methods_detail}. The parameters are set differently due to the differences in target size in each sequence. In practice, we can preset empirical parameters and adjust them according to the feedback results. The visualized output of trajectory extraction is shown in Fig \ref{Track_Extraction_fig}. Since the trajectory extraction method sets the target trajectory to 255 and the background to 0, it is no longer appropriate to use the ROC, SCRG, and BSF as evaluation metrics. We use the TPR and FPR instead. The metrics are shown in Table \ref{Track_Extract_method_metrics}.

From the experimental results, we can see that our trajectory extraction method extracts the target trajectory well and eliminates all the false alarm points. Compared with the traditional threshold method, our method can effectively reduce the false alarm rate. Since our method presets a lower threshold to eliminate a large number of false alarms in the background, our method can also improve the detection rate. Moreover, our trajectory extraction method runs very fast, extracting trajectories in less than 0.3 seconds. The additional time consumption is acceptable relative to the excellent performance.

\begin{figure}[h]
\centering
\includegraphics[width=3in]{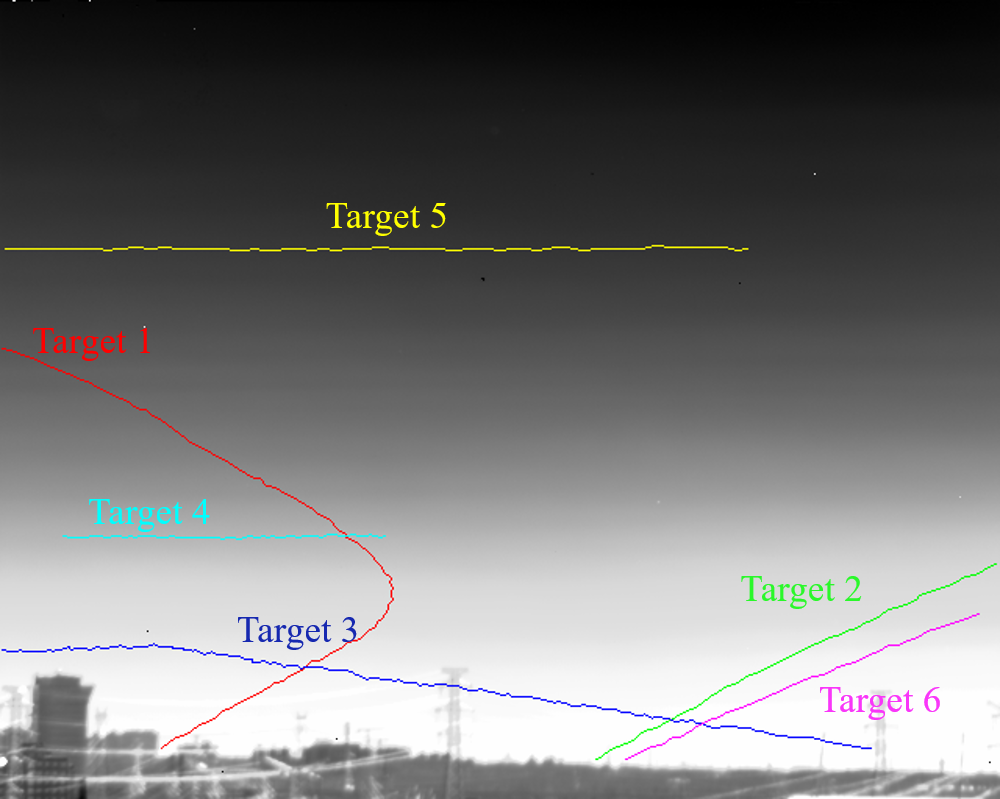}
\caption{Six trajectories formed by six targets.}
\label{MWIR3_fig}
\end{figure}

\begin{figure}[t]
    \centering
    \includegraphics[width=3in]{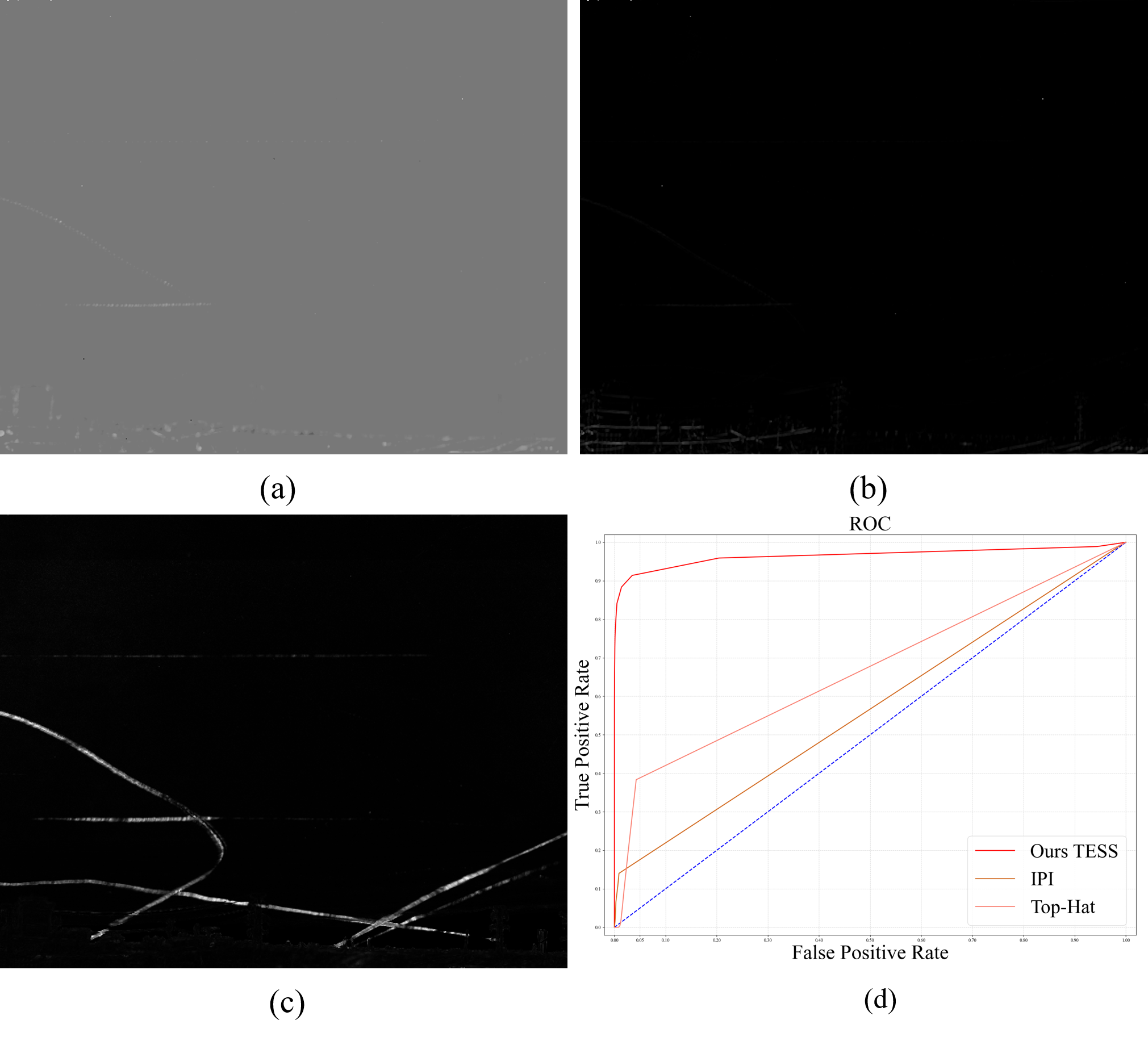}
    \caption{Detection results. (a) Result of IPI. (b) Result of Top Hat. (c) Result of our detection method. (d) ROC of all detection methods.}
    \label{MWIR3_result_fig}
\end{figure}

\begin{figure}[h]
    \centering
    \includegraphics[width=3.5in]{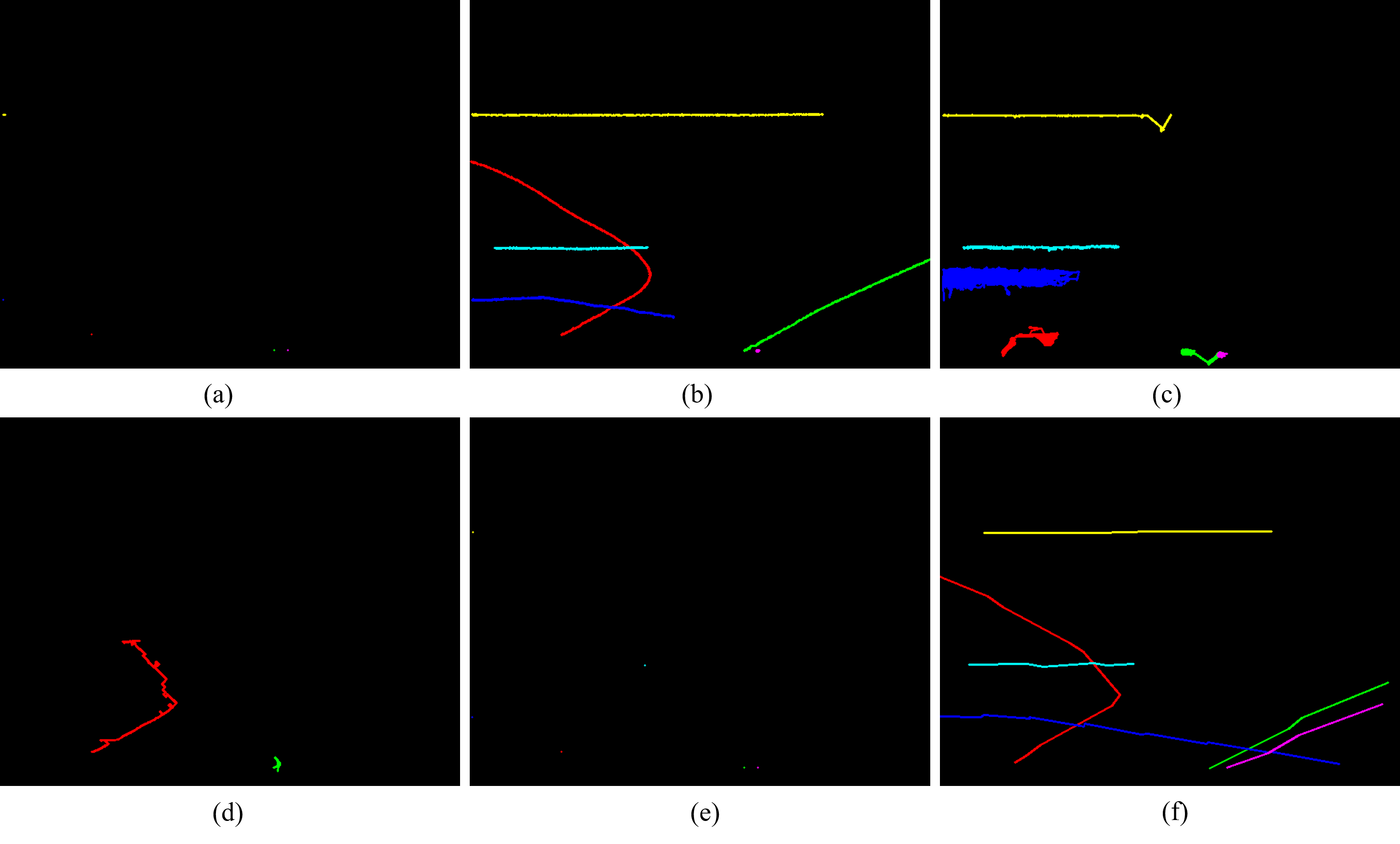}
    \caption{Tracking results. (a) Result of KCF. (b) Result of CSRT. (c) Result of MIL. (d) Result of BOOSTING. (e) Result of MOSSE. (f) Result of our tracking method.}
    \label{MWIR3_track_fig}
\end{figure}

\subsection{Experiments of Tracking Method}
Finally, we conduct experiments on tracking methods. Since the detection result of our method is a segment of the target's trajectory, based on this, we propose a trajectory-based multi-target tracking method. Instead of using the traditional scarce features of a small target on a single frame image, we use the trajectory formed by the target on a multi-frame for tracking, which can effectively solve the tracking problem of multiple weak targets.

\begin{table}[t]
\centering
\caption{AUC and Running Time of Detection Methods.}
\label{AUC_Time_detection}
\begin{tabular}{ccc}
  \hline
  Method & AUC & Running Time (frame/ms)    \\
  \hline
  Ours TESS & 0.9653 & 2.2\\
  Top-Hat\cite{Zeng_2006_design} &0.6683 & 26.7 \\
  IPI\cite{Gao_2013_Infrared} &0.5663 & 6090.87 \\
  \hline
\end{tabular}
\end{table}

\begin{table*}[h]
\centering
\caption{Metrics of Tracking Methods.}
\label{Track_method_metrics}
\begin{tabular}{c|c|c|c|c|c|c|c|c|c|c|c|c}
  \hline
    & \multicolumn{12}{c}{Target}\\
  \hline
    & \multicolumn{2}{c|}{1}& \multicolumn{2}{c|}{2}& \multicolumn{2}{c|}{3}& \multicolumn{2}{c|}{4}& \multicolumn{2}{c|}{5}& \multicolumn{2}{c}{6}\\
\hline
  Method &TPR&FPR&TPR&FPR&TPR&FPR&TPR&FPR&TPR&FPR&TPR&FPR\\
  \hline
  Ours & \textbf{100.00\%}&\textbf{0.00\%}&99.61\%&\textbf{0.00\%}&\textbf{100.00\%}&\textbf{0.00\%}&\textbf{100.00\%}&\textbf{0.00\%}&90.13\%&\textbf{0.00\%}&\textbf{100.00\%}&\textbf{0.00\%}\\
  KCF\cite{Henriques_2015_High-Speed} & 0.47\%&0.00\%&0.78\%&0.00\%&0.36\%&0.00\%&0.00\%&0.00\%&1.26\%&0.00\%&0.88\%&0.00\%\\
  CSRT \cite{Lukezic_2017_Discriminative}& 89.86\%&0.00\%&\textbf{100.00\%}&0.00\%&51.87\%&0.00\%&90.43\%&0.00\%&\textbf{100.00\%}&0.00\%&2.20\%&0.00\%\\
  MIL \cite{Babenko_2011_Robust}& 7.31\%&0.37\%&0.00\%&0.18\%&1.43\%&1.78\%&90.00\%&0.00\%&62.82\%&0.29\%&0.00\%&0.06\%\\
  BOOSTING \cite{Oza_2001_Online}&40.57\%&0.00\%&4.28\%&0.00\%&0.00\%&0.00\%&0.00\%&0.00\%&0.00\%&0.00\%&0.00\%&0.00\%\\
  MOSSE \cite{Bolme_2010_Visual}&0.47\%&0.00\%&0.78\%&0.00\%&0.36\%&0.00\%&1.30\%&0.00\%&0.63\%&0.00\%&0.88\%&0.00\%\\
  
  \hline
\end{tabular}
\end{table*}

\subsubsection{Sequence for Tracking Experiment}
 We used a mid-wave infrared camera to monitor the airspace in sequence 4 and sequence 5 for 90 seconds, during which time six airplanes flew over, producing 4,500 frames of imagery. The target trajectories are shown in Fig.\ref{MWIR3_fig}. Each colored line represents a trajectory formed by a target. The details of this sequence are shown in Table \ref{Detail_Track_Seq_Table}. Of these targets, targets 1, 2, and 6 took off from the airport, targets 4 and 5 had flights over this airspace, and target 3 landed. These targets have lower SCR and smaller sizes and fly at different speeds. In addition, the trajectories between targets 2 and 3 overlap. These increase the difficulty of correct tracking.

\subsubsection{Baseline Methods}
As mentioned in the introduction, traditional single-frame-based small target tracking methods can be divided into two categories. The first is tracking without detection, which tracks the target after manually entering the target position. The second is tracking while detecting, which tracks targets based on the detection result. This method consists of a detector and a tracker. We compare our trajectory-based tracking methods with these two types of tracking methods. In theory, our tracking method belongs to tracking while detection, but we track based on the target trajectory formed by multiple frames instead of a single frame.

Firstly, we conducted experiments on tracking while detection methods. For detector selection, we used the IPI\cite{Gao_2013_Infrared} with the best detection effect and the Top-Hat\cite{Zeng_2006_design} with the fastest speed and then performed tracking method experiments after obtaining the detection results. We analyze whether the detection results and detection speed satisfy the tracking requirements, and if they do, we use the same tracking strategy as in our tracking method. As can be expected, the detector is unable to meet the tracking requirements due to the large resolution and low SCR of targets. Then, we conducted experiments on tracking without detection methods, using KCF\cite{Henriques_2015_High-Speed}, CSRT\cite{Lukezic_2017_Discriminative}, MIL\cite{Babenko_2011_Robust}, BOOSTING\cite{Oza_2001_Online}, and MOSSE\cite{Bolme_2010_Visual} as trackers. The parameters of all tracking methods are shown in Table \ref{tracking_methods_detail}.

\subsubsection{Analysis of Tracking Results}
We first analyze the detection performance of IPI, Top-Hat, and our detection method. The detection results are shown in Fig.\ref{MWIR3_result_fig} and Table\ref{AUC_Time_detection}. It can be seen that IPI and Top-Hat are almost impossible to detect the target and the background contains a lot of clutter that cannot be applied to the tracking method. In addition, their running time is too long, with IPI taking more than 6 seconds to detect an image, which makes them completely inapplicable to real warning systems. Our method can effectively detect the trajectories of all targets and the detection time is 2.2ms, which can meet the tracking requirements. 

The tracking results of all tracking methods are shown in Fig.\ref{MWIR3_track_fig}. It is evident that KCF and MOSSE did not track any targets. The CSRT encountered difficulties in tracking targets 3 and 6. Target 3 was obstructed by a tower during its movement, leading to the tracker losing track of it. In the case of target 6, the low SCR during takeoff resulted in the tracker's failure to extract reliable target features, causing tracking failure right from the start. The MIL can only track targets 4 and 5 with a cleaner background, while the rest of the targets cannot be tracked stably. The BOOSTING was only able to track target 1 and eventually lost it. Our method can track all targets stably. However, for target 5, the detection method cannot effectively detect the trajectory at the end of target 5's motion due to the gradual decrease of the target's SCR, which leads to the trajectory loss. The TPR, FPR, and tracking time for each target is shown in Table \ref{Track_method_metrics}. 

\begin{table}[h]
\centering
\caption{Running Time of Tracking Methods.}
\label{Track_Time}
\begin{tabular}{c|c}
  \hline
    Method    & Running Time (s)  \\
  \hline
  Ours & 12.09 \\
  KCF\cite{Henriques_2015_High-Speed} & 3.09\\
  CSRT \cite{Lukezic_2017_Discriminative} & 133.34 \\
  MIL \cite{Babenko_2011_Robust} & 582.12\\
  BOOSTING \cite{Oza_2001_Online} & 6.31 \\
  MOSSE \cite{Bolme_2010_Visual} & 0.28\\
  
  \hline
\end{tabular}
\end{table}

The running time is shown in Table \ref{Track_Time}. The detection time and tracking time of our method combined is 12.09 seconds, which is significantly faster than the CSRT and MIL methods. The KCF, BOOSTING, and MOSSE methods have too poor tracking performance in spite of their faster tracking speed. Therefore, our method runs the fastest and has the best performance and can be well applied to autonomous warning systems.

\section{Conclusion}

In this article, we have done a lot of work on the problem of small target detection and tracking under low SCR. Firstly, we theoretically analyse the drawbacks of the single-frame detection method and the advantages of the multi-frame detection method, and fully investigate the statistical properties of the target, background and noise components in the ITPs formed by the pixels in the multi-frame detection unit. Based on the differences in statistical properties among the components in the ITP, we propose a detection method based on temporal energy selective scaling to suppress the background and noise components in ITP and amplify the target signal. Extensive experiments demonstrate the high performance of our detection method. Subsequently, to resolve the contradiction between detection rate and false alarm rate brought by the traditional threshold segmentation, we fuse the occurrence matrix and position matrix of the detection results into 3D space, and extracted the target trajectory using 3D Hough transform. Comparing with the threshold segmentation method, our trajectory extraction method can suppress the most false alarms without losing the detection rate. Finally, we study the multi-small target tracking problem. We model the states of the extracted local trajectories, design the cost matching function, and improve the existing tracking strategies, propose a trajectory-based multi-small target tracking method. Experiments with multiple tracking methods show that our method possesses the best tracking performance with the least running time. Our proposed target detection, trajectory extraction and target tracking methods all have the advantages of short running time and good performance, and can well support early warning and monitoring systems. However, our method also has shortcomings. If the target moves very fast or the detector frame rate is too low, the target's motion process cannot be fully recorded, resulting in too little temporal information and poor detection performance. For camera-moving scenarios, the ITP of a pixel varies so drastically that it is impossible to estimate the statistical properties, leading to the inapplicability of our method. In addition, the performance of the tracking method is heavily dependent on the detection results. In the future, we will further use the target's spatial features in combination with the temporal features in ITP to increase the range of applications of our methods. We will also improve the tracking method by incorporating the idea of tracking without detection method to make it capable of autonomous tracking.

\section*{Acknowledgments}
This work was partly supported by the Youth Innovation Promotion Association, Grant NO. E1213A02, and the Key Research Program of Frontier Sciences, CAS, Grant NO. 22E0223301.
\bibliographystyle{ieeetrans.bst}
\bibliography{reference.bib}

\end{document}